\documentclass[sigconf]{acmart}

\AtBeginDocument{%
  \providecommand\BibTeX{{%
    \normalfont B\kern-0.5em{\scshape i\kern-0.25em b}\kern-0.8em\TeX}}}

\newcommand{\nameshort}[1]{EfficientGEBD}
\newcommand{\nameshorth}[1]{xxxxHead}

\newcommand{\namebasic}[1]{BasicGEBD}

\usepackage{multirow}

\usepackage{xcolor}
\usepackage{colortbl}

\definecolor{mygray}{gray}{.9}

\newcommand{\Yb}{\mathbf{Y}}

\begin{document}

\title{Rethinking the Architecture Design for Efficient Generic Event Boundary Detection}

\author{Ziwei Zheng}
\affiliation{%
 \institution{Xi'an Jiaotong University}
 \country{China}}
\email{ziwei.zheng@stu.xjtu.edu.cn}

\author{Zechuan Zhang}
\affiliation{%
 \institution{Xi'an Jiaotong University}
 \country{China}}
\email{2213712891@stu.xjtu.edu.cn}

\author{Yulin Wang}
\affiliation{%
 \institution{Tsinghua University}
 \country{China}}
\email{wang-yl19@mails.tsinghua.edu.cn}

\author{Shiji Song}
\affiliation{%
 \institution{Tsinghua University}
 \country{China}}
\email{shijis@mail.tsinghua.edu.cn}

\author{Gao Huang}
\affiliation{%
 \institution{Tsinghua University}
 \country{China}}
\email{gaohuang@tsinghua.edu.cn}

\author{Le Yang}
\authornote{Corresponding author.}
\affiliation{%
 \institution{Xi'an Jiaotong University}
 \country{China}}
\email{yangle15@xjtu.edu.cn}



\begin{abstract}

Generic event boundary detection (GEBD), inspired by human visual cognitive behaviors of consistently segmenting videos into meaningful temporal chunks, finds utility in various applications such as video editing and. In this paper, we demonstrate that SOTA GEBD models often prioritize final performance over model complexity, resulting in low inference speed and hindering efficient deployment in real-world scenarios. We contribute to addressing this challenge by experimentally reexamining the architecture of GEBD models and uncovering several surprising findings. Firstly, we reveal that a concise GEBD baseline model already achieves promising performance without any sophisticated design. Secondly, we find that the widely applied image-domain backbones in GEBD models can contain plenty of architecture redundancy, motivating us to gradually ``modernize'' each component to enhance efficiency. Thirdly, we show that the GEBD models using image-domain backbones conducting the spatiotemporal learning in a spatial-then-temporal greedy manner can suffer from a distraction issue, which might be the inefficient villain for GEBD. Using a video-domain backbone to jointly conduct spatiotemporal modeling is an effective solution for this issue. The outcome of our exploration is a family of GEBD models, named EfficientGEBD, significantly outperforms the previous SOTA methods by up to 1.7\% performance gain and 280\% speedup under the same backbone. Our research prompts the community to design modern GEBD methods with the consideration of model complexity, particularly in resource-aware applications. The code is available at \url{https://github.com/Ziwei-Zheng/EfficientGEBD}.

\end{abstract}

\ccsdesc[500]{Computing methodologies~Video segmentation}
\ccsdesc[500]{Computing methodologies~Activity recognition and understanding}

\keywords{Generic event boundary detection, Video understanding}


\maketitle
\begin{figure}[t]
    \centering
    \includegraphics[width=0.99\linewidth]{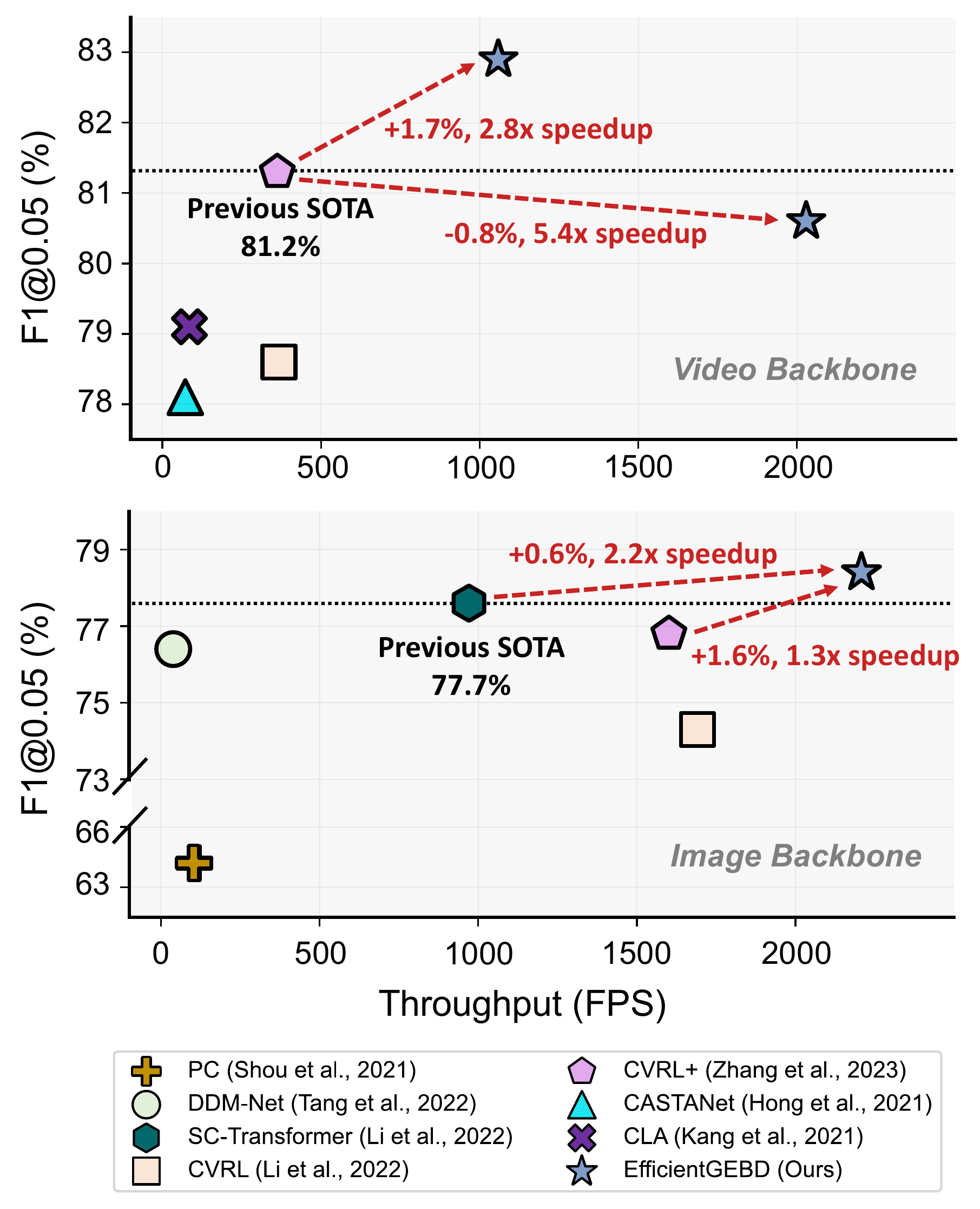}
    \vspace{-10pt}
    \caption{The throughput \textit{vs.} F1 score of GEBD methods on Kinetics-GEBD~\cite{shou2021generic}.}
    \label{fig1}
    \vspace{-10pt}
\end{figure}

\section{Introduction}

Video understanding has gained significant traction in multimedia fields recently~\cite{feng2023refinetad,mizufune2023margin,zheng2019relation}. Motivated by cognitive science showing human naturally tends to parse long videos into meaningful segments for subsequent comprehending~\cite{radvansky2011event}, researchers have proposed the Generic Event Boundary Detection task (GEBD)~\cite{shou2021generic} that aims at detecting the moments in videos as generic and taxonomy-free event boundaries. Generally, the development of GEBD task will be valuable in immediately supporting applications like video editing~\cite{choi2015automated} and summarization~\cite{he2019unsupervised}, and more importantly, spurring progress in long-form video, where GEBD can be implemented as the first step towards segmenting video into units for further reasoning and understanding.

The potential value of GEBD has promoted the development of the benchmark competition, such as LOVEU21-23~\cite{loveu22,loveu21,loveu23}. However, the requirements of such competitions tend to encourage the models to achieve high performance without considering model complexity. From Figure~\ref{fig1}, we see that some of the existing GEBD methods~\cite{kang2021winning,tang2022progressive,shou2021generic,hong2021generic} indeed suffer from the efficiency issue and have low throughput. On one hand, efficiency itself is supposed to be an important evaluation metric for any GEBD method. On the other hand, as GEBD can serve as a pre-processing step, its high latency will surely lead to the inefficiency of understanding of long-form videos.


This paper rethinks the design paradigm of the deep network-based GEBD model and is intended to improve its efficiency. 
We surprisingly find that a basic GEBD model containing the most concise design can already achieve comparable detection performance (F1 score of 77.1\%) with the soft-label training techniques in~\cite{li2022structured,zhang2023local}. We call this basic GEBD model as \namebasic{}. Considering that early GEBD methods~\cite{shou2021generic,tang2022progressive} do not apply soft-label during training, we imply a portion of the detection performance differences between existing GEBD methods may be due to these training techniques.


We then use the \namebasic{} as the starting point to gradually ``modernize'' each component of \namebasic{} towards a highly efficient GEBD model. Our investigation of the backbone reveals that the commonly applied ResNet50~\cite{he2016deep} contains plenty of redundancy and increasing the capacity of the backbone in \namebasic{} is no benefit to GEBD performance. This motivates us to reduce the size of the backbone to improve efficiency. We then reexamine the rest components and find that lightweight designs can effectively achieve temporal modeling for GEBD. Specifically, we build the encoder based on difference maps~\cite{tang2022progressive} capturing the local relation and use a small convolution network to process the similarity matrix to extract the global information. Then we further propose to use a cross-attention module to fuse global-local information for final predictions. We show that each modified component is effective yet lightweight compared to previous methods. Finally, we obtain a family of GEBD models named \nameshort{}, which can achieve a new SOTA result (78.3\%) on Kinetics-GEBD while with 2.2$\times$ speedup than the previous SOTA methods~\cite{li2022structured} using the same backbone (Figure~\ref{fig1}).

Then we naturally think about why using high-capacity backbone models does not benefit the final GEBD performance. Our results show that using image-domain backbones conducting the spatiotemporal learning in a spatial-then-temporal greedy manner can lead to a \textit{Distraction} issue, which can distract the attention of the backbone from the true boundary-related objects. Such an issue can be due to the absence of temporal modeling ability of the image-domain backbone and therefore will be effectively addressed by implementing a video-domain backbone to jointly conduct spatiotemporal modeling for GEBD, which further boosts the performance of \nameshort{} by a large margin.

We hope the new observations and discussions can challenge some common designs and existing evaluation metrics in GEBD tasks. First, we suggest considering efficiency as an important metric for evaluation to ensure the applicability of GEBD models. Second, using small image-domain backbones is sufficient for GEBD models which improves the efficiency with high detection performance. Third, developing a GEBD model directly based on the video-domain backbone is suggested for future works. Our contributions can be summarized as follows:
\vspace{-5pt}
\begin{itemize}
    \item We introduce a strong baseline model for GEBD tasks, \namebasic{}, which achieve high performance without any sophisticated designs.
    \item By detailed studying each component of \namebasic{}, we obtain a family of GEBD models, named \nameshort{}, achieving SOTA performance with high inference speed.
    \item Implementing video-domain backbone for spatiotemporal modeling can significantly boost the performance of \nameshort{} (82.9\%) and even \namebasic{} (82.5\%).
    \item Extensive experiments and studies on the Kinetics-GEBD \cite{shou2021generic} and TAPOS \cite{shao2020intra} datasets provide new experimental evidence as well as demonstrate the effectiveness and efficiency of our work on GEBD tasks.
\end{itemize}

\vspace{-5pt}
\section{Methodology}
\label{sec3}

\begin{figure}[t]
    \centering
    \includegraphics[width=1.02\linewidth]{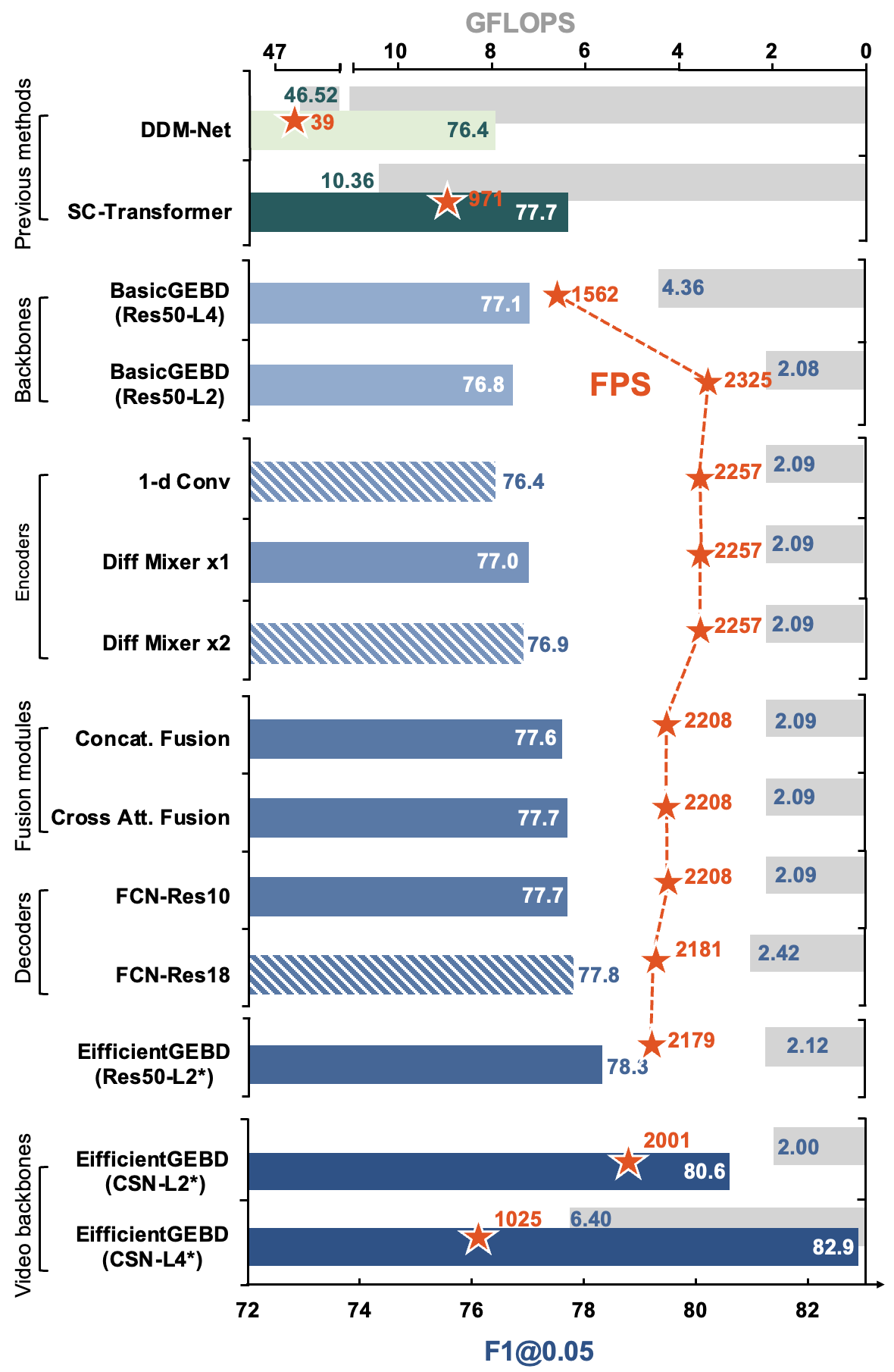}
    \vspace{-15pt}
    \caption{We modernize the proposed \namebasic{} towards the design of an efficient GEBD model. The colored bars are the F1@0.05 scores of models and the gray bars depict the GFLOPs. A hatched bar means the modification is not adopted. The orange stars mean FPS. In the end, our \nameshort{} with ResNet50-L2* can outperform the previous SOTA method (SC-Transformer~\cite{li2022structured}), and can be further obviously improved by using CSN backbone~\cite{tran2019video}.}
    \label{fig_roadmap}
    \vspace{-10pt}
\end{figure}

In this section, we provide a trajectory going from a basic GEBD model to an efficient GEBD model. We first build a baseline model, named \namebasic{}, as a starting point by abstracting existing popular GEBD methods~\cite{tang2022progressive,li2022structured,kang2022uboco,shou2021generic}. By reexamining each component of \namebasic{}, we propose to reduce the surplus computational costs and improve the model efficiency, resulting in a new family of efficient GEBD models, named \nameshort{}. Moreover, by investigating the possible reason why the larger size of the backbone model turns into surplus costs, we are motivated to jointly conduct spatiotemporal modeling in the backbone by using a video domain network when building GEBD models. In Figure~\ref{fig_roadmap}, we show the procedure and the results we are able to achieve with each step of the ``model modernization''. All models in Section~\ref{sec3} are trained and evaluated on Kinetics-GEBD~\cite{shou2021generic}, a challenging and well-known GEBD dataset, which will be introduced in Section~\ref{sec:exp}.

\textbf{Training settings.} We uniformly sample 100 frames (i.e., $T=100$) from each video as inputs. The length of a video clip is set to 17, where the model is designed to recognize whether the median frame is the boundary or not. We apply the training settings in~\cite{li2022structured,zhang2023local} that use Gaussian Smoothing and the whole model is trained end-to-end for 15 epochs with Adam~\cite{kingma2014adam}. The learning rate is set as 1e-2, being divided by 10 at the 6th and 8th epochs. We will use this fixed training recipe with the same hyper-parameters throughout this section. We first briefly introduce the Gaussian Smoothing.

According to~\cite{li2022structured}, Gaussian Smoothing proposes to smooth the sparse one-hot labels, $\Yb \in \{0,1\} ^{T}$, with a Gaussian kernel with the width of $\sigma$ to generate soft labels $\widetilde{\Yb} \in [0,1]^{T}$, where $T$ is the length of video. The smoothed label can provide more boundary information and is effective in improving detection performance. We use $\sigma=1$ for all our experiments.

\textbf{Evaluation.} The Relative Distance (Rel.Dis.) representing the error between the detected and ground truth timestamps divided by the length of the action instance, is used for evaluation. Following~\cite{shou2021generic,loveu21,loveu22,loveu23}, we report the F1 scores with the Rel.Dis. of 0.05 in this section. Moreover, theoretical (GFLOPs) and practical (FPS) speeds are used for efficiency evaluation. The throughput is tested using the maximal batch size for each model running on a NVIDIA RTX 4090 GPU with mixed precision.



\subsection{A baseline model for GEBD: \namebasic{}}

Although GEBD methods can also be designed in unsupervised or self supervised settings, in this paper, we mainly focus on the GEBD models that treat the boundary detection tasks as supervised video clip binary classification tasks in an end-to-end manner. Other types, such as detection-based GEBD models with off-line feature extraction (Temporal Perceiver~\cite{tan2023temporal}), are not included in our study. SBoCo~\cite{kang2022uboco}, DDM-Net~\cite{tang2022progressive}, and SC-Transformer~\cite{li2022structured} (SC-Trans.) are selected as the representative supervised GEBD networks. By studying their architectures, we show that these models follow the five components design paradigm: (1) The backbone for feature extraction; (2) The encoder for temporal modeling; (3) The similarity map (Sim. Map); (4) The decoder processing the similarity map; (5) The feature fusion module. These components are summarized in Table~\ref{tab:base}. Generally, these methods conduct the spatiotemporal modeling in a greedy step-by-step manner, where the backbone is first used to extract the spatial representations and then the temporal modeling is conducted by the subsequent modules. 

\begin{figure}[htp]
    \centering
    \includegraphics[width=0.99\linewidth]{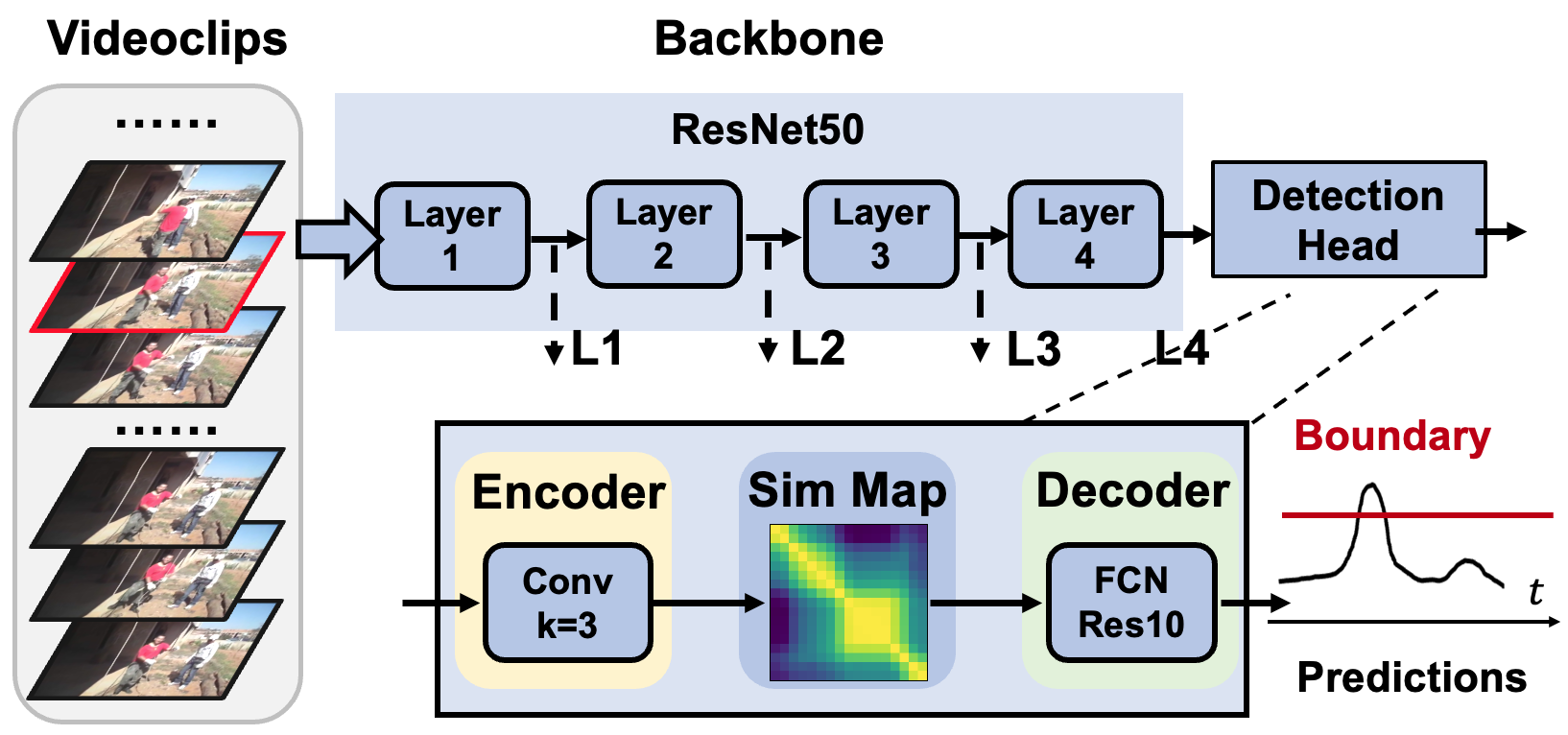}
    \caption{The architecture of \namebasic{}.}
    \label{fig_base}
\end{figure}

We use the most concise design for each component build the baseline: (1) The widely used ResNet50~\cite{he2016deep} pre-trained on ImageNet~\cite{deng2009imagenet} is applied as the backbone. (2) A 1-d Conv layer, consisting of BN~\cite{ioffe2015batch}, Conv (Kernel=3) and ReLU, is implemented as the encoder. (3) The cosine similarity (CosSim.) is used to generate the similarity matrix. (4) A fully convolutional network (FCN), which normally consists of a mini ResNet10\footnote{A ResNet10 has 4 residual layers, where each layer contains 1 residual block, consisting of two convolution layers with residual connection.}~\cite{zhang2023local,tang2022progressive,li2022end} as the decoder. As only DDM-Net applies the fusion module, we do not use it in our baseline model. The obtained architecture is shown in Figure~\ref{fig_base} and we call this basic GEBD model \namebasic{} in our research.

Surprisingly, we found that the performance of the \namebasic{} can be up to 77.1\% in the term of F1@0.05 (shown in Table~\ref{tab:exp_kgebd}). It is amazing that such a concise model can achieve high detection results without any additionally sophisticated designs as well as also show high efficiency in terms of both FLOPs and FPS (shown in Figure~\ref{fig_roadmap} and Table~\ref{tab:base}). From the results, we conclude that apart from the design of the GEBD network, the training procedure also affects the ultimate performance of GEBD models. Therefore, we infer that the superior performance of some modern GEBD models actually benefits not only from the sophisticated architecture design but also from the advanced training techniques.

\paragraph{We then use the \namebasic{} as the start point to explore a more efficient model for GEBD tasks.}

\begin{table*}[t]
\centering
\caption{The architectures of three representative GEBD methods and the propose models in this paper. }
\label{tab:base}
\resizebox{\linewidth}{!}{
\begin{tabular}{llllllcc}
\toprule
Mehtods & Backbone (FLOPs) & Encoder (FLOPs) & Sim. Map & Decoder (FLOPs) & Fusion (FLOPs) & F1@0.05 & FLOPs\\
\midrule
\multirow{2}{*}{SboCo~\cite{kang2022uboco}$\dag$} & \multirow{2}{*}{ResNet50} & \multirow{2}{*}{1d-Conv}& CosSim. or &  ResNet +  & \multirow{2}{*}{-} & \multirow{2}{*}{73.2} & \multirow{2}{*}{163.5G} \\
 & & &L2-Sim. & Transformer \\ \midrule
DDM-Net~\cite{tang2022progressive}  & ResNet50 (4.10G) & Differences (0.0G) & L2-Sim. & FCN (0.25G) & Progressive Att. (1.02G) & 76.4 & 46.52G\\
\midrule
SC-Trans.~\cite{li2022structured} & ResNet50 (4.10G) & Transformer (60M) & CosSim. & FCN (5.92G) & - & 77.7 & 10.36G \\
\midrule\midrule
\namebasic{}     & ResNet50-L4 (4.10G) & 1d-Conv (0.04M)& CosSim. & FCN (0.25G) & - & 77.1 & 4.36G  \\
\midrule
\nameshort{} & ResNet50-L2* (1.82G)  & DiffMixer (17.9M) & CosSim. & FCN (0.25G) & Cross Att. (0.8M) & 78.3 & 2.12G \\
\nameshort{} & CSN-L2* (1.72G)  & DiffMixer (17.9M) & CosSim. & FCN (0.25G) & Cross Att. (0.8M) & 80.6 & 2.00G \\
\nameshort{} & CSN-L4* (6.09G)  & DiffMixer (26.8M) & CosSim. & FCN (0.25G) & Cross Att. (0.8M) & 82.9 & 6.40G \\
\bottomrule
\multicolumn{8}{l}{\small $\dag$ The FLOPs of each component can not be calculated since the official code is not available. The overall FLOPs are referred from~\cite{gothe2023self}. }
\end{tabular}
}
\end{table*}

\subsection{Exploring for \nameshort{}}
\label{sec_backbone}

Our exploration from \namebasic{} to \nameshort{} is by gradually “modernizing” the \namebasic{} from four aspects: 1) backbone network; 2) encoder; 3) decoder; 4) fusion module, where the backbone is for spatial modeling, and the rest are designed for temporal modeling. In Figure~\ref{fig_roadmap}, we show this procedure and the results we are able to achieve with each step of the model modernization.

\subsubsection{Backbone network}
\label{sec221}
From Table~\ref{tab:base}, we see that the backbones (ResNet50) generally have a large computational costs for both \namebasic{} and other GEBD methods, which motivates us to investigate how the size of the backbone model can affect the performance of a GEBD model.

\begin{figure}[htp]
    \centering
     \vspace{-5pt}
    \includegraphics[width=0.99\linewidth]{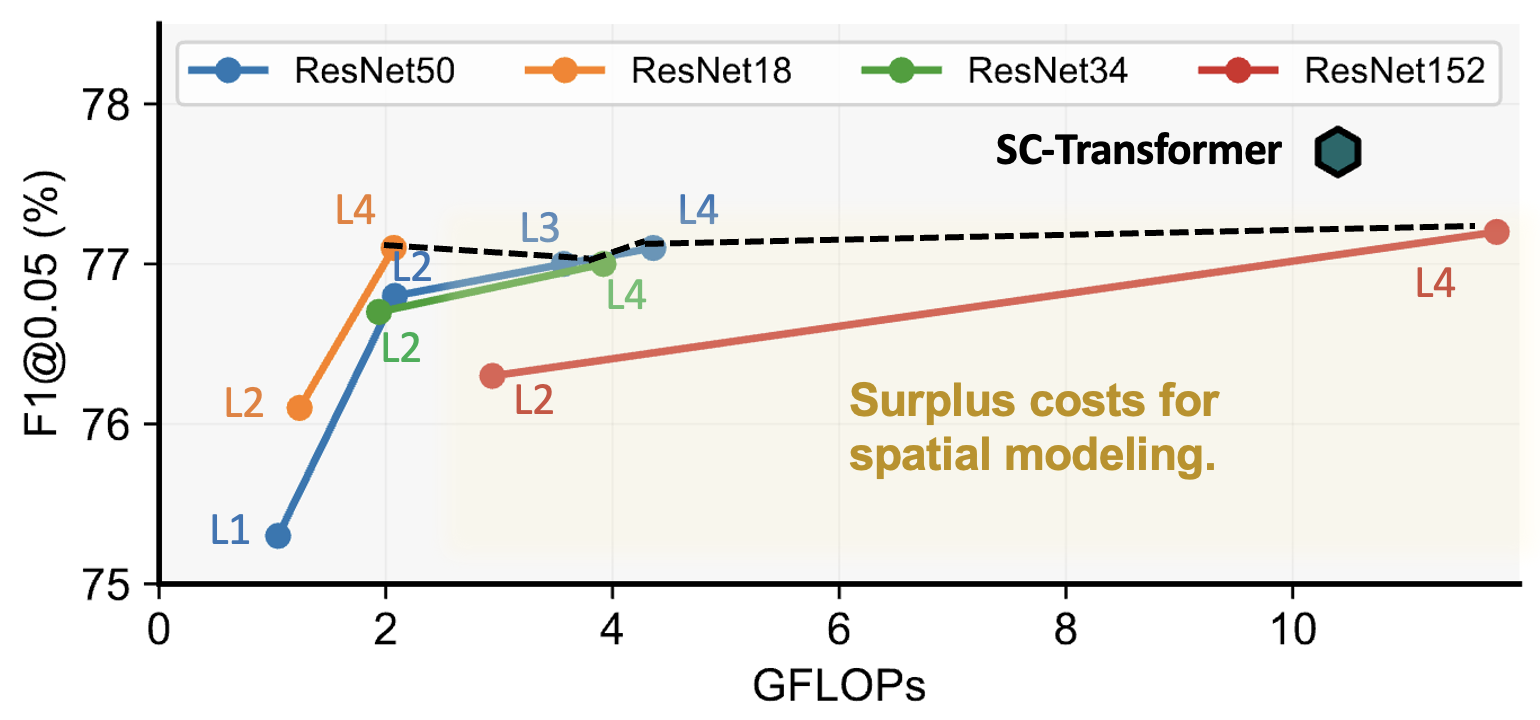}
    \vspace{-10pt}
    \caption{The GFLOPs \textit{v.s} F1 score of \namebasic{} with different sizes of ResNets as the backbone.}
    \label{fig_baseres}
    \vspace{-5pt}
\end{figure}

This motivates us to test the \namebasic{} with different sizes of ResNets (Figure~\ref{fig_baseres}). From the results, we see that using a larger backbone network does not lead to better performance: the \namebasic{} with ResNet18 can already achieve the detection accuracy compared to that with ResNet50, indicating the extra computational costs of ResNet50 can be redundant, which can be further confirmed by the results of \namebasic{} with ResNet152. This finding contradicts the general common sense in other vision tasks, such as object detection~\cite{lin2017focal,lin2017feature}, where using a larger backbone model normally brings obvious performance improvements. Considering that the ResNets are used for spatial modeling in these GEBD models, we hypothesize that there exist tons of redundancy for spatial modeling when using ResNet50 or larger models as the backbone in \namebasic{}. Such a hypothesis is also confirmed by the high performance achieved by the \namebasic{} when we attach the detection head at early layers of ResNet50 (E.g, ResNet50-L2), where the model achieves valuable detection performance (76.8\%), which is already superior to most of the existing GEBD methods (see the results in Table~\ref{tab:exp_kgebd}). Also, the \namebasic{} with ResNet50-L2 of 2.08 GFLOPs is over 45\% smaller than the original \namebasic{}, approximately 5 times smaller than SC-Transformer~\cite{li2022structured} (10.36 GFLOPs) and over 20 times smaller than DDM-Net~\cite{tang2022progressive} (40 GFLOPs) (shown in Figure~\ref{fig_roadmap}), leading to the fast throughput of the modified GEBD model (2,325 FPS). We also found that the performance of \namebasic{} can drastically drop with too small a capacity of backbone, indicating that the capacity of the backbone for GEBD should be large enough for spatial modeling.



Due to the observed redundancy in spatial modeling when we use ResNet50 as the backbone, we can improve the efficiency of our model with barely performance loss by reducing the size of the backbone. For a fair comparison with the previous researches~\cite{li2022structured,tang2022progressive} using ResNet50, we use ResNet50-L2 to conduct the following examinations. However, our findings also hold for small-size backbones, such as ResNet18, which will be shown in the experiments.

\textit{Then, we will use the first two layers of the backbone network (ResNet50-L2) to build the efficient GEBD model.}



\vspace{-10pt}
\subsubsection{Temporal modeling ability in encoder}
\label{sec222}

We then try to increase the temporal modeling ability for boundary detection. We show that compared to the self-attention-based encoder~\cite{li2022structured}, the concise design in~\cite{tang2022progressive} introducing difference maps for temporal modeling can achieve better performance. Such a difference map also meets our intuition in boundary detection: To detect these motion-related boundaries, motion information plays a principal role in perceiving temporal variations and can be effectively modeled by using feature differences at different timestamps.







\begin{figure*}[htp]
    \centering
    \includegraphics[width=0.99\linewidth]{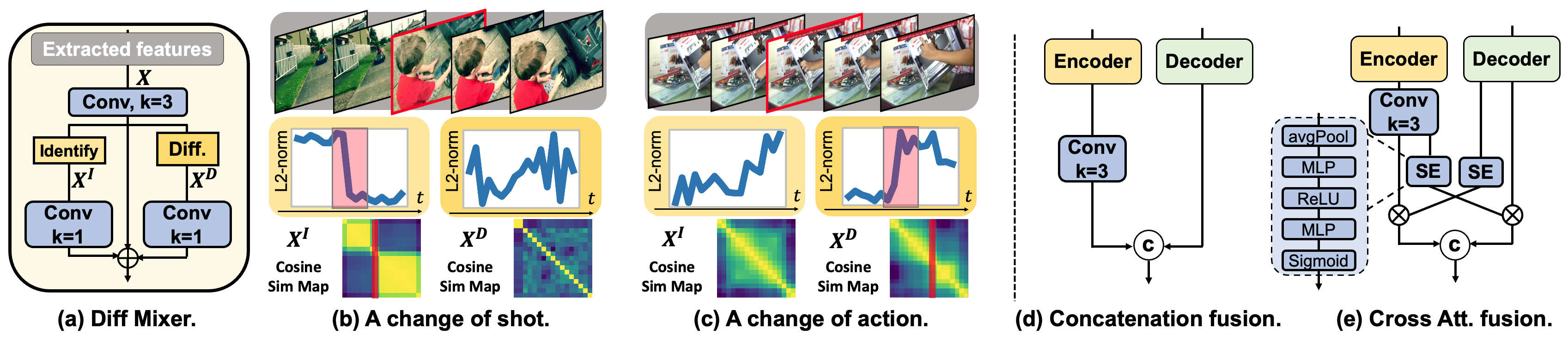}
    \vspace{-10pt}
    \caption{The illustrations of the encoder (a) and the fusion module (d,e). In (b,c), we calculate the L2-norm and the cosine similarity map of the features at different timestamps to see whether the discriminative boundary features can be captured.}
    \label{fig_encoder}
    \vspace{-10pt}
\end{figure*}

Figure~\ref{fig_encoder} (a) shows the architecture of the proposed difference mixer (Diff Mixer) encoder. Different from~\cite {tang2022progressive}, we use both difference features, $\mathbf{X}^D$, and the original features, $\mathbf{X}^I$, for temporal modeling. Given the extracted features $\mathbf{X} \in \mathcal{R}^{T\times C}$, where $T$ is the temporal length and the $C$ is the number of channels, the difference can be calculated by $\mathbf{X}^D_t = \mathbf{X}_{t} - \mathbf{X}_{t+1}$, $t=1,2,...,T$. And we further pad $\mathbf{X}_{1}$ at the beginning of $\mathbf{X}^D$ to guarantee that $\mathbf{X}^D$ and $\mathbf{X}$ have the same dimension. Then $\mathbf{X}^D$ and $\mathbf{X}^I$ will be processed by a conv layer (BN-Conv-ReLU) with the kernel size of 1.

Such a design enables the encoder to effectively model the local relationship among different video frames: shot changes and motion-related boundaries can be effectively identified by the original $\mathbf{X}^{I}$ and difference $\mathbf{X}^{D}$ features, respectively, which can be illustrated by Figure~\ref{fig_encoder} (b,c). We plot the variants of the features norm and similarity maps for $\mathbf{X}^{I}$ and $\mathbf{X}^{D}$. The shot changes can be obviously identified with the extracted features without using differences. However, for the action changes, temporal difference seems a more important cue for boundary identifying.

As depicted in Figure~\ref{fig_roadmap}, the implementation of Diff Mixer can boost the performance of our model to 77.0\% while approximately maintaining the inference speed. We also test using the Marco design of Transformers~\cite{li2022structured} as the encoder, and observe that there are no obvious improvements in the performance. Increasing the number of Diff Mixers also does not benefit the GEBD performance (the F1 score drops to 76.9\% when we use two Diff Mixers). More studies can be found in supplementary materials.

\textit{We then use one Diff Mixer as the encoder.}

\vspace{-5pt}
\subsubsection{Feature fusion module}
\label{sec223}

Feature fusion which introduces the features at final predictions, is achieved by Progressive Att. in~\cite{tang2022progressive} to improve the performance. However, the computational costs of Progressive Att. can be up to 1.02G. In this section, we investigate how to conduct such a feature fusion mechanism with a lightweight design to improve the performance of GEBD.


Two fusion modules shown in Figure~\ref{fig_encoder} (d,e) are tested, where the fusion can be achieved by either directly concatenating two features or using cross attention. For concatenation, the features from the encoder will be firstly processed by a Conv layer (BN-Conv-ReLU) with a kernel size of 3, and then concatenated to the features from the decoder. For the cross attention (Cross Att.) in Figure~\ref{fig_encoder} (e), two squeeze-and-excitation (SE)~\cite{hu2018squeeze} are used to generate the weights for re-weighting the resultant features from each branch.

The Figure~\ref{fig_roadmap} shows that two fusion mechanisms can boost the performance to 77.6\% and 77.7\%, respectively. The proposed fusion model is significantly smaller than that in~\cite{tang2022progressive} (shown in Table~\ref{tab:base}), and therefore barely affects the efficiency. So far, the model can be comparable to the previous SOTA SC-Transformer~\cite{li2022structured} with ResNet50 backbone. However, the modified \namebasic{} with Cross Att. can run over 100\% faster than SC-Transformer in terms of FPS.

We further indicate that such improvements can be due to the fusion of global-local temporal information. On one hand, the decoder can gather global information by using the FCN to process the similarity matrix of the whole video clip. Therefore, the features from the decoder can be viewed as the global representations of the event boundary. On the other hand, the difference features from Diff Mixer measure the relationship for the adjacent frames (local information), which potentially captures motion cues in the temporal dimension (shown in Figure~\ref{fig_encoder} (b,c)). Introducing them in the final predictions can immediately improve the identification of some event boundaries, and therefore benefit whole GEBD tasks. 

\textit{We then use Cross Att. fusion in our model.}

\subsubsection{The FCN in decoder}
\label{sec224}
The implementation of the similarity maps as well as the 2-d FCN decoder can be seen as a sign of recent outperforming supervised GEBD methods. The FCN normally consists of a mini ResNet (such as ResNet10 in~\cite{tang2022progressive}, or ResNet18 in~\cite{zhang2023local,li2022end}) or its variants (a 4-layer fully convolutional network in~\cite{li2022structured}). Such a design is important to achieve accurate detection performance for GEBD models. For example, if we apply the ``old fashion'' design in~\cite{shou2021generic} without the similarity map and the FCN decoder (the features from encoder will be directly used for final predictions), the performance of \namebasic{} will drastically drop to 64.6\% even with the DiffMixer.

From Figure~\ref{fig_roadmap} we see that the model with ResNet18 as the decoder is only 0.1\% higher than that using ResNet10 (77.7\% \textit{v.s} 77.8\%), indicating that increasing the size of the FCN does not necessarily lead to better performance. The applied decoder only has 0.25G FLOPs, much smaller than the decoder in~\cite{li2022structured} (5.92G). Therefore, we indicate that the large decoder in the SC-Transformer might contain too much redundancy, limiting its efficiency.

\textit{We then use a ResNet10 as the decoder for the final model.}

\vspace{-5pt}
\subsubsection{\nameshort{}}
We have finished our first “playthrough” and obtain the proposed GEBD model. Moreover, following~\cite{tang2022progressive}, we introduce both the features from layer-1 and layer-2 (L1 and L2) and concatenate them as the inputs for the encoder, which further boost the performance of the model from 77.7\% to 78.3\%, while hardly affects the inference speed. Here, we use ResNet50-L2* to denote that both the features from layers 1 and 2 are used. To this end, we have discovered the efficient GEBD model, named \nameshort{}, that achieves SOTA performance using ResNet50 as the backbone on the Kinetics-GEBD dataset. Remarkably, our \nameshort{} achieve 0.6\% higher performance while having 2.2$\times$ speedup compared to the previous SOTA GEBD method, SC-Transformer~\cite{li2022structured}. 

Although we have explored ways to reduce the surplus computational costs for spatial modeling to build \nameshort{}, the inefficiency issue of the backbone network is still like a dark cloud on the horizon of building high-performance GEBD models. Indeed, the results in Figure~\ref{fig_baseres} show that scaling up the backbone to ResNet152 does not obviously help in performance improvement of \nameshort{}. To achieve higher detection performance, recent studies~\cite{zhang2023local,tan2023temporal} and the winner solutions of GEBD competition~\cite{hong2021generic,hong2022sc,sun2023mae} have proposed to implement deep models for video domain to build GEBD methods, such as Two-streams networks (TSN)~\cite{feichtenhofer2016convolutional} and channel-separated video network (CSN)~\cite{tran2019video}. The applications of CSNs built based on ResNet152 significantly boost the performance of the GEBD models in~\cite{zhang2023local,li2022end}. As the used CSN is also built based on ResNet152, we are interested in the reason why using video backbone brings obvious improvements, which is further investigated in the following.

\vspace{-10pt}
\subsection{Distraction issue in GEBD}
\label{sec23}

So far, all previously examined GEBD models conduct the spatial and temporal modeling in a greedy learning manner, where an image-domain backbone, such as a ResNet, is firstly used to extract the spatial features, and then the subsequent modules are applied to explore the temporal information for event boundary detection. 

\begin{figure}[htp]
    \centering
    \vspace{-10pt}
    \includegraphics[width=0.49\textwidth]{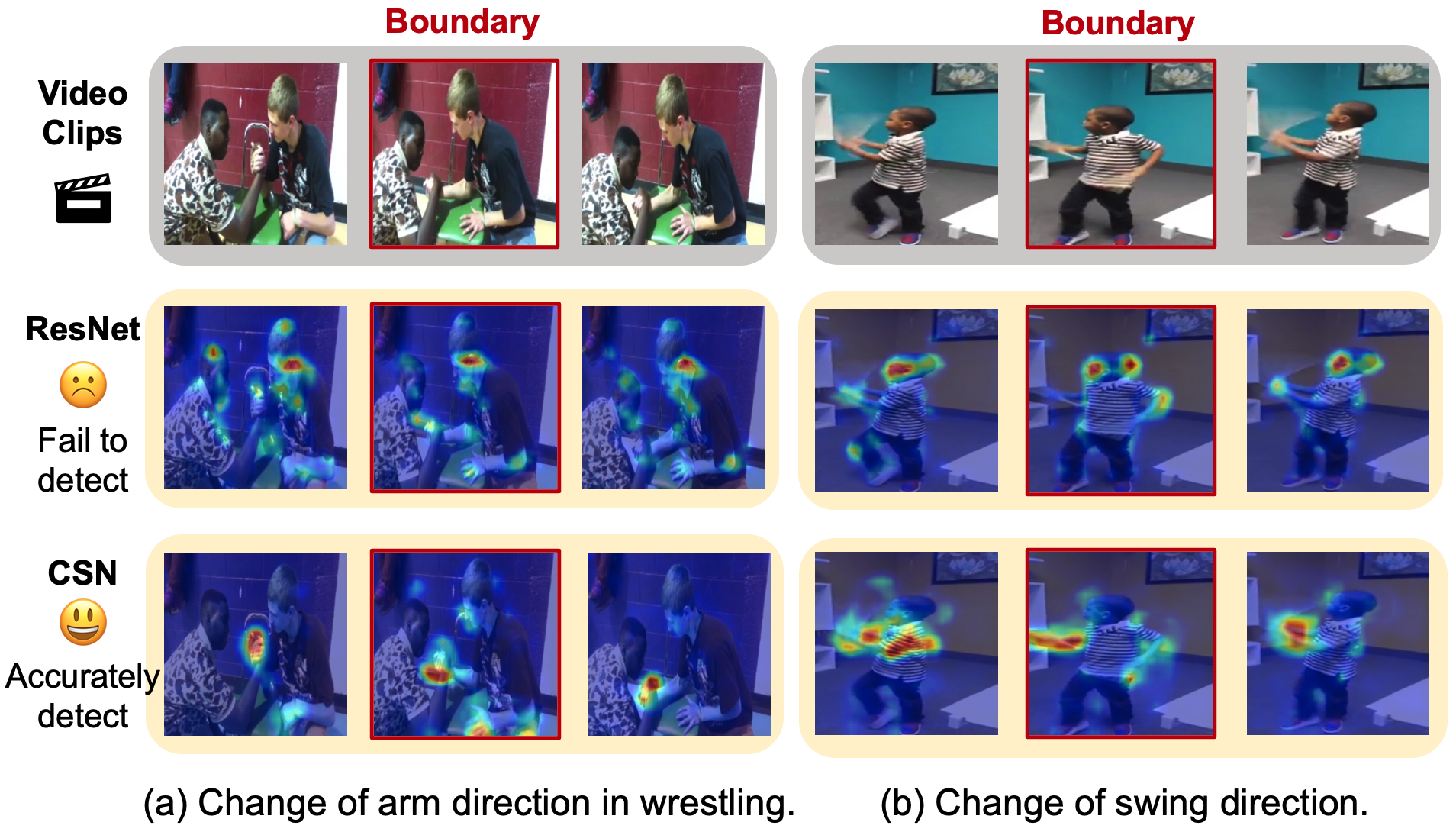}
    \vspace{-15pt}
    \caption{The activations captured by GradCAM++~\cite{chattopadhay2018grad} using ResNet~\cite{he2016deep} and CSN~\cite{tran2019video} as the backbones for GEBD models. The median frame is the boundary frame.}
    \label{fig_cam}
    \vspace{-10pt}
\end{figure}

However, we hypothesize that conducting the spatiotemporal learning in such a greedy way can lead to several inefficiency issues in GEBD. As the image domain backbones are usually designed to identify the main objects in an image, learning the spatial features without the guide from temporal information can result in the attention of the backbone distracting from the objects most related to the boundaries, and getting stuck in some areas containing the main objects in each frame. We refer to such an issue as \textit{distraction issue}. Figure~\ref{fig_cam} illustrates the distraction issue. For instance, the event boundary in Figure~\ref{fig_cam} (a) is defined by the changes of action during arm wrestling. The high activations of the ResNet backbone get stuck in the spatial areas that contain the head of the person in the center of the frame. With the arm wrestling-related spatial areas missing features extracted from the backbone, the subsequent modules will have difficulties conducting the following temporal modeling, resulting in the failure detection of this boundary. Therefore, we indicate the distraction issue may be the real villain for the inefficiency when we use image backbones to build GEBD models.



However, we see that high activations of using channel-separated video network (CSN)~\cite{tran2019video} surround the arm wrestling-related spatial areas. Actually, spatial and temporal learning can complement each other during feature learning. On one hand, temporal information can be important in guiding the backbone to focus on boundary-related spatial representations. On the other hand, the learned spatial features can be used to explore the temporal variants, which are important cues for boundary detection. Therefore, instead of learning spatiotemporal features in a step-by-step manner, a more reasonable way is to jointly extract the spatiotemporal features by using video-domain backbones, which effectively avoids the aforementioned distraction issue and improves the GEBD performance.


To this end, CSN~\cite{tran2019video}, pre-trained on video action recognition datasets, IG65M~\cite{ghadiyaram2019large}, is implemented as the backbone to improve \namebasic{} and \nameshort{}. By using CSN-L4 as the backbone, the \namebasic{} surprisingly achieves the performance of 82.5\%, which outperforms all previous published researches~\cite{zhang2023local,tan2023temporal,li2022end}. Moreover, the \nameshort{} with CSN-L2 and CSN-L4 can achieve the performance of 80.4\% and 82.0\%, respectively (these results are reported in supplementary materials). Compared to the scenario that introducing additional model capacity does not bring obvious GEBD performance using an image-domain backbone, the superiority of CSN-L4 compared to CSN-L2 indicates that the backbone redundancy does exist when we use video-domain backbone, which demonstrates the importance of conducting spatiotemporal modeling in the backbone to build high-performance GEBD models. 

As depicted in Figure~\ref{fig_roadmap}, \nameshort{} with CSN-L2* and CSN-L4* achieve the performance of 80.6\% and 82.9\%. However, our simple yet effective models are still highly efficient, \nameshort{} with CSN-L2* achieving a throughput of over 2,000, which is over 4$\times$ faster than previous SOTA result in~\cite{zhang2023local}. When using CSN-L4*, we can achieve a new SOTA result of 82.9\% on Kinetics-GEBD with over 1,000 FPS, which is 180\% faster than~\cite{zhang2023local}. These encouraging findings prompt us to rethink the correctness of using an image domain backbone when building GEBD models. 

\textbf{Remark.} In some video tasks, image-based models are effective and efficient, which seems to conflict with our results. Such a conflict can be due to the differences between the tasks: For tasks such as video instance segmentation, the models are required to achieve spatial locating with the assistance of temporal relations. In an extreme case, video instance segmentation can be roughly achieved by image segmentation models segmenting frame-by-frame. We indicate that image models can be effective in these video tasks. However, boundary detection in GEBD can only be achieved based on analyzing the temporal information among a series of frames with strong temporal modeling ability, which can not be realized with a single frame input.

\section{Experiment}
\label{sec:exp}

\subsection{Experimental Settings}


\subsubsection{Datasets.} The evaluations are based on the frequently used Kinetics-GEBD~\cite{shou2021generic}, TAPOS~\cite{shao2020intra} and SoccernetV2~\cite{deliege2021soccernet}. Kinetics-GEBD contains 54,691 videos randomly selected from Kinetics-400 \cite{carreira2017quo}, which are labeled with 1,290,000 generic event temporal boundaries. The distribution of videos across training, validation, and testing sets in Kinetics-GEBD is nearly uniform, maintaining a ratio close to 1:1:1. Each video is annotated by five annotators, resulting in an average of approximately 4.77 boundaries per video. Since the annotations for the test set are not publicly accessible, we conducted training on the training set and subsequently evaluated model performance using the validation set. In line with the methodology outlined in \cite{shou2021generic}, we adapt TAPOS by concealing each action instance's label and conducting experiments based on this modified dataset. SoccernetV2 is a large-scale soccer video dataset. Following~\cite{tan2023temporal}, we select 200 games annotated with 158,493 shot boundaries for the task of camera shot boundary detection. The annotated games are divided into training, validation and testing set of 120, 40
and 40 recordings, respectively.

\subsubsection{Implementation Details.} The implementation of Kinetics-GEBD is provided in Section~\ref{sec_backbone}. As the videos in TAPOS have a large variety of duration over instances, we split the instances without overlapping and sample 100 frames by keeping a similar FPS to Kinetics-GEBD following~\cite{tan2023temporal}. The scores of sub-instances are merged to generate the final prediction. The whole model is trained end-to-end for 30 epochs with Adam \cite{kingma2014adam} and a base learning rate of 2e-2, which will be divided by 10 at the 6th, 8th, and 15th epochs, respectively. Other settings are the same as these for Kinetics-GEBD. For SoccernetV2, we first temporally down-sample the videos to 5 FPS and then extract the offline features using the backbone. Then we split the features into a series of 20s non-overlapping clips as inputs. The model without the backbone will be trained for 50 epochs with the base learning rate of 1e-3.


\subsection{Empirical Evaluations on Kinetics-GEBD}

\begin{table}[!t]
\centering
\caption{Comparisons in terms o F1 score (\%) on Kinetics-GEBD with Rel.Dis. threshold from 0.05 to 0.5.
}
\vspace{-10pt}
\label{tab:exp_kgebd}
\resizebox{1.0\linewidth}{!}{
\begin{tabular}{l|l|c|cccc}
\toprule
\multirow{2}{*}{Method} & \multirow{2}{*}{Backbone} & \multicolumn{5}{c}{F1 @ Rel. Dis.}   \\ \cline{3-7} 
                    &    & 0.05          & 0.1             & 0.3                   & 0.5           & avg           \\ \midrule
BMN \cite{lin2019bmn}         &     Res50       & 18.6          & 20.4     & 23.0  & 24.1          & 22.3          \\

BMN-StartEnd \cite{lin2019bmn}     &  Res50     & 49.1          & 58.9          & 66.8          & 68.3          & 64.0          \\
TCN-TAPOS \cite{lea2016segmental}     &    Res50      & 46.4          & 56.0          & 65.9          & 68.7          & 62.7          \\
TCN \cite{lea2016segmental}         &   Res50         & 58.8          & 65.7          & 70.3          & 71.2          & 68.5          \\
PC \cite{shou2021generic}          &    Res50        & 62.5          & 75.8          & 85.3          & 87.0          & 81.7          \\
PC+OF \cite{shou2021generic}          &    Res50        & 64.6         & 77.6          & 86.4          & 87.9          & 83.0          \\
SBoCo \cite{kang2022uboco}      &    Res50   & 73.2          & -             & -             & -             & 86.6          \\
Temporal Per. \cite{tan2023temporal}  &  Res50  & 74.8          & 82.8          & 87.9          & 89.2          & 86.0          \\
CVRL \cite{li2022end}        &   Res50         & 74.3          & 83.0           & 88.6          & 89.8          & 86.5          \\
CVRL+ \cite{zhang2023local}  & Res50&76.8  &84.8 &89.6 &90.6 & 87.7\\ 
DDM-Net \cite{tang2022progressive}        &   Res50      & 76.4          & 84.3          & 89.2          & 90.2          & 87.3          \\
SC-Transformer \cite{li2022structured}   &  Res50     & 77.7          & 84.9          & 90.0          & 91.1          & 88.1          \\ 
\rowcolor{mygray} \textbf{\namebasic{}} &  Res50 & 76.8  &83.4   &88.5   &89.6   &86.6\\
\rowcolor{mygray} \textbf{\nameshort{}} &  Res50-L2* & \textbf{78.3}        & \textbf{85.1}       & \textbf{90.1}    & \textbf{91.3}   & \textbf{88.3}
\\\midrule
SBoCo \cite{kang2022uboco}      &    TSN   & 78.7         & -             & -             & -             & 89.2          \\
CLA~\cite{kang2021winning}      &    TSN   & 79.1         & -             & -             & -             & -          \\
CASTANet~\cite{hong2021generic} &    CSN   & 78.1         & -             & -             & -             & -    \\
CVRL \cite{li2022end}        &     CSN       & 78.6          & -           & -         & -          & -        \\
CVRL+ \cite{zhang2023local} & CSN  &81.2 &- &- &- &-\\
\rowcolor{mygray}\textbf{\namebasic{}} & CSN  & 82.5   &87.7     &91.9      &92.8   &90.4    \\
\rowcolor{mygray}\textbf{\nameshort{}} & CSN  & \textbf{82.9}      & \textbf{88.2}      & \textbf{92.2}   & \textbf{93.2} & \textbf{90.8} \\

\bottomrule
\end{tabular}
}\vspace{-10pt}
\end{table}

\begin{figure}[htp]
    \centering
    \includegraphics[width=0.99\linewidth]{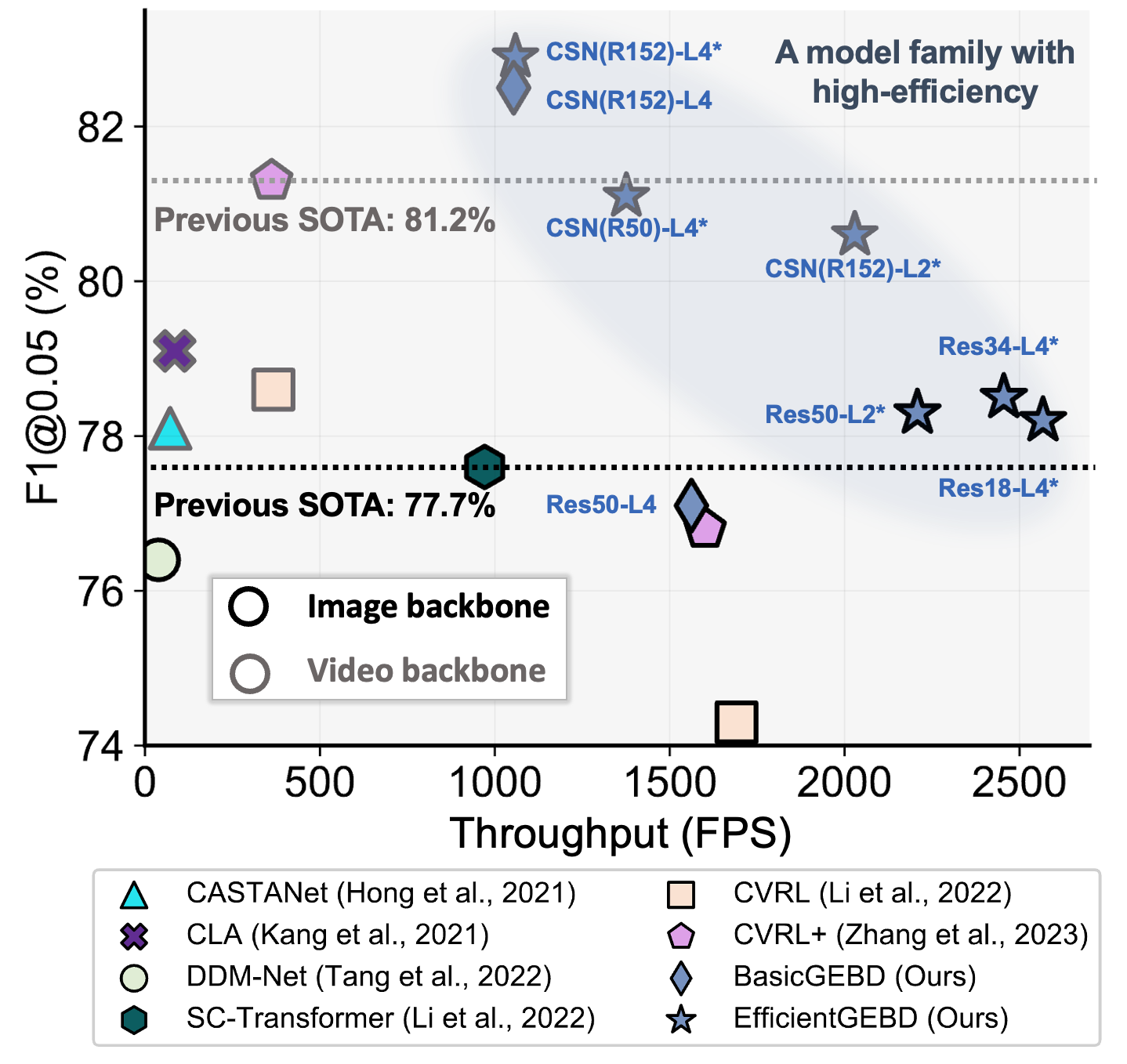}
    \vspace{-10pt}
    \caption{The FPS \textit{v.s} F1 score with evaluated GEBD methods on Kinetics-GEBD. FPS is measured using NVIDIA RTX 4090.}
    \label{fig_kgebdall}
    \vspace{-10pt}
\end{figure}

Table~\ref{tab:exp_kgebd} and Figure~\ref{fig_kgebdall} show the results on the Kinetics-GEBD validation set, with Rel.Dis. threshold from 0.05 to 0.5. Overall, we see that our methods achieve promising performance compared to previous methods in different Rel. Dis.. Moreover, we find that \nameshort{} is superior to previous GEBD methods under the most stringent Rel. Dis. constraint (0.05), indicating the more accurate detection ability of our method. When using the ResNet50 as the backbone, \nameshort{} competes SC-Transformer~\cite{li2022structured} with 0.6\% higher performance (78.3\% \textit{v.s} 77.7\%) and 2.2$\times$ speedup (2208 FPS \textit{v.s} 971 FPS). \nameshort{} achieves new SOTA performance of 82.9\% when using CSN as backbone model, which is over 1.7\% higher and 2.8$\times$ speedup than the CVRL+ in ~\cite{zhang2023local}. We also find that although the \namebasic{} achieves fair detection performance with image-domain backbone, the \namebasic{} with CSN can achieve the detection performance of 82.5\%, which is also superior to CVRL+, indicating the importance of conducting spatiotemporal modeling in the backbone models as we stated in Section~\ref{sec23}. The higher performance of using CSN(R50)-L4* (81.1\%) also confirms our findings. In Figure~\ref{fig_kgebdall}, we show that the family of evaluated \nameshort{}s all achieve high efficiency in the experiments which demonstrate the effectiveness and efficiency of our design.


\subsection{Results on TAPOS}
We also conduct experiments on the TAPOS \cite{shao2020intra} dataset in Table \ref{tab:exp_tapos}. The proposed \nameshort{} also achieves the SOTA performance in terms of F1@0.05 (63.1\%) and average (74.8\%), respectively, which further proves the strong generalizability of our methods. More results and analyses on TAPOS are provided in supplementary materials. This verified the effectiveness and robustness of our method in different scenes.

\begin{table}[t]
\centering
\caption{Comparison with others in terms of F1 score (\%) on TAPOS with Rel.Dis. threshold from 0.05 to 0.5.}
\vspace{-10pt}
\label{tab:exp_tapos}
\resizebox{\linewidth}{!}{
\begin{tabular}{l|c|cccc}
\toprule
\multirow{2}{*}{Method} & \multicolumn{5}{c}{F1 @ Rel. Dis.}                                                                                 \\ \cline{2-6} 
                        & 0.05                     & 0.1                      & 0.3                      & 0.5                      & avg                      \\ \midrule
ISBA \cite{ding2018weakly}                   & 10.6                     & 17.0                     & 32.6                     & 39.6                     & 30.2                     \\
TCN \cite{lea2016segmental}                    & 23.7                     & 31.2                     & 34.4                     & 34.8                     & 64.0                     \\
CTM \cite{huang2016connectionist}                    & 24.4                     & 31.2                     & 36.9                     & 38.5                     & 35.0                     \\
TransParser \cite{shao2020intra}            & 28.9                     & 38.1                     & 51.4                     & 54.5                     & 47.4                     \\
PC \cite{shou2021generic}                     & 52.2                     & 59.5                    & 66.5                     & 68.3                     & 64.2                     \\
Temporal Perceiver \cite{tan2023temporal}     & 55.2 & 66.3 & 76.5 & 78.8 & 73.2 \\
DDM-Net \cite{tang2022progressive}                & 60.4                     & 68.1                     & 75.3                     & 76.7                     & 72.8                     \\
SC-Transformer \cite{li2022structured}         & 61.8 & 69.4 & 76.7 & 78.0 & 74.2 \\ 
\rowcolor{mygray} \textbf{BasicGEBD} (Res50-L4)  & \textbf{60.0}            & \textbf{66.6}            & \textbf{73.1}            & \textbf{74.8}            & \textbf{71.0}            \\

\rowcolor{mygray} \textbf{EfficientGEBD} (Res50-L3*)  & \textbf{62.6}            & \textbf{70.1}            & \textbf{77.2}            & \textbf{78.4}            & \textbf{74.7}            \\
\rowcolor{mygray} \textbf{EfficientGEBD} (Res50-L4*)  & \textbf{63.1}            & \textbf{70.5}            & \textbf{77.4}            & \textbf{78.6}            & \textbf{74.8}            \\ \bottomrule
\end{tabular}
}
\vspace{-5pt}
\end{table}

\subsection{Results on SoccernetV2}
We further implement the proposed \nameshort{} by the two-stage detection manner for long video (>30 mins) boundary detection: we use a pre-trained backbone for feature extraction and then apply the extracted features for boundary detection. The results in Tab.~\ref{tab:exp_soccer} show that the \nameshort{} is also suitable for processing long videos in the two-stage manner and can achieve good performance on shot-level action spotting tasks. 
\begin{table}[t]
\centering
\caption{Comparison for shot-level detection with other methods on SoccernetV2 using 0.3 Rel.Dis.}
\vspace{-10pt}
\label{tab:exp_soccer}
\resizebox{0.95\linewidth}{!}{
\begin{tabular}{l|l|ccc}
\toprule
Method & Backbone & Precision & Recall & F1 score \\ \midrule
Histogram~\cite{sv} & RAW & 87.1 & 65.2 & 75.6 \\
SC-Transformer~\cite{li2022structured} & ResNet152-off & 93.6 & 92.2& 92.9\\
\midrule
\rowcolor{mygray} \textbf{EfficientGEBD} & ResNet152-off  &  94.4 & \textbf{94.8} & 94.6           \\
\rowcolor{mygray} \textbf{EfficientGEBD} & CSN-off  &  \textbf{95.6} & 94.1 & \textbf{94.8}           \\ \bottomrule
\end{tabular}
}
\vspace{-17pt}
\end{table}

\subsection{Visualization}

\begin{figure}[htp]
    \centering
    \includegraphics[width=1.05\linewidth]{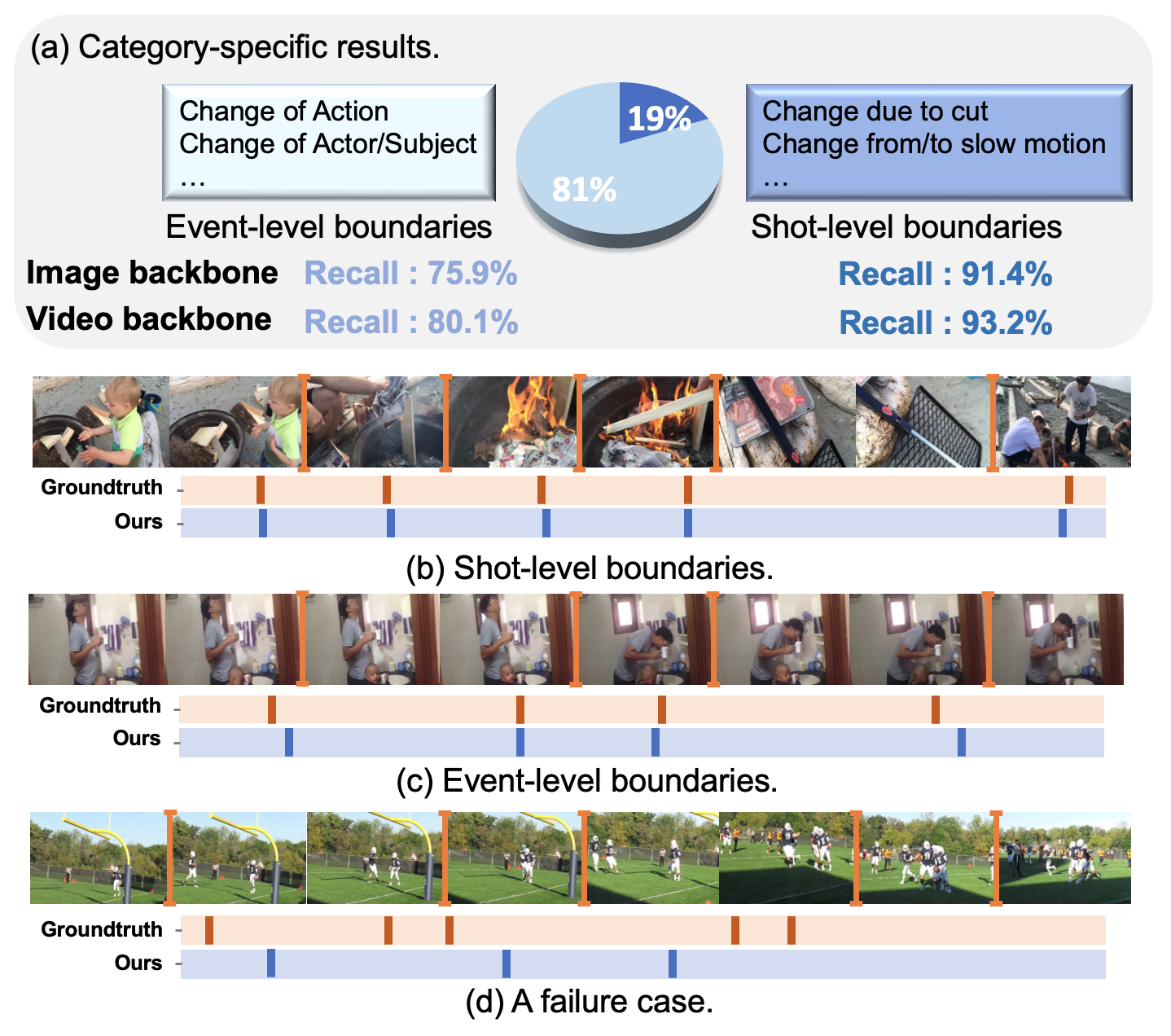}
    \vspace{-15pt}
    \caption{Detection results on Kinetics-GEBD for (a) category-specific detection recalls. (b) shot-level changes, (c) event-level changes and (d) a failure case. }
    \label{fig_vis}
    \vspace{-10pt}
\end{figure}

The generic event boundaries can be specifically categorized into shot- (19\%) and event-level (81\%) boundaries. In this experiment, we further evaluate the performance of our methods in detecting different kinds of boundaries, where the pseudo recall\footnote{Since the GEBD models only generate the boundary predictions regardless of their specific categories, the True Positive might not be true.} for each category is calculated and shown in Figure~\ref{fig_vis} (a). We see that shot-level boundaries can be mostly effectively detected by the GEBD models. These event-level boundaries that have the largest number (81\%) are the samples that indeed make up the bulk of the miss-detection in GEBD tasks. The results also meet our hypothesis stated in Section~\ref{sec23}, that the image backbone is more likely to suffer from the distraction issue when detecting event-level boundaries, resulting in a low detection recall (75.9\%). While using a video backbone effectively addresses the issue and can achieve an obvious detection improvement (80.1\%) for these event-level boundaries.

We further provide qualitative results for boundary detection on Kinetics-GEBD in Figure~\ref{fig_vis}, from which we see that most predictions of our method are accurate. We also find that our method struggles with the detection when there are multiple small objects in the videos(Figure~\ref{fig_vis} (d)). This meets our intuition since the changes of multiple objects can all be viewed as boundaries, which increase the complexity of detecting them.

\section{Related Work}
\label{sec:related}

In video understanding fields, temporal detection tasks normally involve identifying clip-level instances from within untrimmed videos, such as shot boundary detection~\cite{gygli2018ridiculously,souvcek2019transnet,tang2018fast}, temporal action segmentation~\cite{aakur2019perceptual,farha2019ms,lei2018temporal}, and temporal action localization~\cite{zhang2022actionformer,nag2023post,cheng2022tallformer,shi2023tridet}. Initially introduced in~\cite{shou2021generic}, GEBD targets the localization of taxonomy-free moments, mirroring human perception of event boundaries according to recent cognitive science. These boundaries serve as crucial cues for a deeper understanding of long-form videos. Most existing research models the GEBD tasks as a binary classification problem for the input video clips. A line of research for GEBD tasks adopts a similar approach as described in~\cite{shou2021generic} to partition lengthy videos into adjacent overlapping snippets, treating them as independent samples~\cite{hong2021generic,kang2021winning,gothe2023self}. DDM-Net~\cite{tang2022progressive} further proposes to characterize the motion pattern with dense difference maps. There are also works that conduct the entire video as a single input stream with continuous predictions~\cite{li2022structured, huynh2023generic} and achieve higher efficiency. Moreover, Temporal Perceiver~\cite{tan2023temporal} builds the offline framework and achieves GEBD using a detection-based method. 

Recent researchers have proposed to build the GEBD models in self-supervised or unsupervised manners. \cite{kang2022uboco} proposes a recursive parsing algorithm based on the temporal self-similarity matrix to enhance local modeling. \cite{rai2023motion} builds a motion-aware GEBD method that uses a differentiable motion feature learning module to tackle spatial and temporal diversities. Furthermore, a parameter-free unsupervised GEBD detector is also proposed in~\cite{gothe2024s}, which conducts GEBD using optical flow without deep architectures.

It is more recent studies have also proposed to focus more on improving the efficiency of the GEBD models. In~\cite{gothe2023self}, a light-weighted GEBD detector is built based on transformer decoders. Moreover, an end-to-end compressed video representation learning (CVRL) for GEBD is proposed in~\cite{li2020learning,zhang2023local}. As recent studies~\cite{zheng2023dynamic,wu2018compressed} have shown that using compressed video streams can speedup the inference, we believe that our method can also achieve higher efficiency when modified to allow the compressed video stream as input.

\section{Conclusion}

As an important pre-processing technique for long-form video understanding, the inefficiency of GEBD methods can largely affect the efficiency of long-form video processing, which prompts us to rethink the architecture design for efficient GEBD. In some ways, the performance and efficiency of the proposed \namebasic{} and \nameshort{} are surprising while the model itself is not new, where most of the designs have been introduced in previous GEBD studies~\cite{li2022structured,tang2022progressive,shou2021generic}. We also believe that combining other parallel techniques, such as self-supervised training~\cite{kang2022uboco,sun2023mae} or using compressed videos as inputs~\cite{li2022end,zhang2023local}, can further improve the performance of our method. We hope that the new results of this study will bring new intuitions for GEBD architecture design and new evaluation metrics that consider the model efficiency for GEBD tasks in video understanding fields.

\bibliographystyle{ACM-Reference-Format}
\bibliography{sample-base}


\begin{thebibliography}{56}


\ifx \showCODEN    \undefined \def \showCODEN     #1{\unskip}     \fi
\ifx \showDOI      \undefined \def \showDOI       #1{#1}\fi
\ifx \showISBNx    \undefined \def \showISBNx     #1{\unskip}     \fi
\ifx \showISBNxiii \undefined \def \showISBNxiii  #1{\unskip}     \fi
\ifx \showISSN     \undefined \def \showISSN      #1{\unskip}     \fi
\ifx \showLCCN     \undefined \def \showLCCN      #1{\unskip}     \fi
\ifx \shownote     \undefined \def \shownote      #1{#1}          \fi
\ifx \showarticletitle \undefined \def \showarticletitle #1{#1}   \fi
\ifx \showURL      \undefined \def \showURL       {\relax}        \fi
\providecommand\bibfield[2]{#2}
\providecommand\bibinfo[2]{#2}
\providecommand\natexlab[1]{#1}
\providecommand\showeprint[2][]{arXiv:#2}

\bibitem[Aakur and Sarkar(2019)]%
        {aakur2019perceptual}
\bibfield{author}{\bibinfo{person}{Sathyanarayanan~N Aakur} {and} \bibinfo{person}{Sudeep Sarkar}.} \bibinfo{year}{2019}\natexlab{}.
\newblock \showarticletitle{A perceptual prediction framework for self supervised event segmentation}. In \bibinfo{booktitle}{\emph{IEEE Conf. Comput. Vis. Pattern Recog.}} \bibinfo{pages}{1197--1206}.
\newblock


\bibitem[Carreira and Zisserman(2017)]%
        {carreira2017quo}
\bibfield{author}{\bibinfo{person}{Joao Carreira} {and} \bibinfo{person}{Andrew Zisserman}.} \bibinfo{year}{2017}\natexlab{}.
\newblock \showarticletitle{Quo vadis, action recognition? a new model and the kinetics dataset}. In \bibinfo{booktitle}{\emph{IEEE Conf. Comput. Vis. Pattern Recog.}} \bibinfo{pages}{6299--6308}.
\newblock


\bibitem[Chattopadhay et~al\mbox{.}(2018)]%
        {chattopadhay2018grad}
\bibfield{author}{\bibinfo{person}{Aditya Chattopadhay}, \bibinfo{person}{Anirban Sarkar}, \bibinfo{person}{Prantik Howlader}, {and} \bibinfo{person}{Vineeth~N Balasubramanian}.} \bibinfo{year}{2018}\natexlab{}.
\newblock \showarticletitle{Grad-cam++: Generalized gradient-based visual explanations for deep convolutional networks}. In \bibinfo{booktitle}{\emph{IEEE Winter Conf. App. Comput. Vis.}} IEEE, \bibinfo{pages}{839--847}.
\newblock


\bibitem[Cheng and Bertasius(2022)]%
        {cheng2022tallformer}
\bibfield{author}{\bibinfo{person}{Feng Cheng} {and} \bibinfo{person}{Gedas Bertasius}.} \bibinfo{year}{2022}\natexlab{}.
\newblock \showarticletitle{Tallformer: Temporal action localization with a long-memory transformer}. In \bibinfo{booktitle}{\emph{Eur. Conf. Comput. Vis.}} Springer, \bibinfo{pages}{503--521}.
\newblock


\bibitem[Choi and Lee(2015)]%
        {choi2015automated}
\bibfield{author}{\bibinfo{person}{Jun-Ho Choi} {and} \bibinfo{person}{Jong-Seok Lee}.} \bibinfo{year}{2015}\natexlab{}.
\newblock \showarticletitle{Automated video editing for aesthetic quality improvement}. In \bibinfo{booktitle}{\emph{ACM Int. Conf. Multimedia}}. \bibinfo{pages}{1003--1006}.
\newblock


\bibitem[Deliege et~al\mbox{.}(2021)]%
        {deliege2021soccernet}
\bibfield{author}{\bibinfo{person}{Adrien Deliege}, \bibinfo{person}{Anthony Cioppa}, \bibinfo{person}{Silvio Giancola}, \bibinfo{person}{Meisam~J Seikavandi}, \bibinfo{person}{Jacob~V Dueholm}, \bibinfo{person}{Kamal Nasrollahi}, \bibinfo{person}{Bernard Ghanem}, \bibinfo{person}{Thomas~B Moeslund}, {and} \bibinfo{person}{Marc Van~Droogenbroeck}.} \bibinfo{year}{2021}\natexlab{}.
\newblock \showarticletitle{Soccernet-v2: A dataset and benchmarks for holistic understanding of broadcast soccer videos}. In \bibinfo{booktitle}{\emph{IEEE Conf. Comput. Vis. Pattern Recog.}}
\newblock


\bibitem[Deng et~al\mbox{.}(2009)]%
        {deng2009imagenet}
\bibfield{author}{\bibinfo{person}{Jia Deng}, \bibinfo{person}{Wei Dong}, \bibinfo{person}{Richard Socher}, \bibinfo{person}{Li-Jia Li}, \bibinfo{person}{Kai Li}, {and} \bibinfo{person}{Li Fei-Fei}.} \bibinfo{year}{2009}\natexlab{}.
\newblock \showarticletitle{Imagenet: A large-scale hierarchical image database}. In \bibinfo{booktitle}{\emph{IEEE Conf. Comput. Vis. Pattern Recog.}} Ieee, \bibinfo{pages}{248--255}.
\newblock


\bibitem[Developers(2015)]%
        {sv}
\bibfield{author}{\bibinfo{person}{S.-V. Developers}.} \bibinfo{year}{2015}\natexlab{}.
\newblock \showarticletitle{Scikit-video: Video processing in python}. In \bibinfo{booktitle}{\emph{https://github.com/scikit-video/scikit-video}}.
\newblock


\bibitem[Ding and Xu(2018)]%
        {ding2018weakly}
\bibfield{author}{\bibinfo{person}{Li Ding} {and} \bibinfo{person}{Chenliang Xu}.} \bibinfo{year}{2018}\natexlab{}.
\newblock \showarticletitle{Weakly-supervised action segmentation with iterative soft boundary assignment}. In \bibinfo{booktitle}{\emph{IEEE Conf. Comput. Vis. Pattern Recog.}} \bibinfo{pages}{6508--6516}.
\newblock


\bibitem[Farha and Gall(2019)]%
        {farha2019ms}
\bibfield{author}{\bibinfo{person}{Yazan~Abu Farha} {and} \bibinfo{person}{Jurgen Gall}.} \bibinfo{year}{2019}\natexlab{}.
\newblock \showarticletitle{Ms-tcn: Multi-stage temporal convolutional network for action segmentation}. In \bibinfo{booktitle}{\emph{IEEE Conf. Comput. Vis. Pattern Recog.}} \bibinfo{pages}{3575--3584}.
\newblock


\bibitem[Feichtenhofer et~al\mbox{.}(2016)]%
        {feichtenhofer2016convolutional}
\bibfield{author}{\bibinfo{person}{Christoph Feichtenhofer}, \bibinfo{person}{Axel Pinz}, {and} \bibinfo{person}{Andrew Zisserman}.} \bibinfo{year}{2016}\natexlab{}.
\newblock \showarticletitle{Convolutional two-stream network fusion for video action recognition}. In \bibinfo{booktitle}{\emph{IEEE Conf. Comput. Vis. Pattern Recog.}} \bibinfo{pages}{1933--1941}.
\newblock


\bibitem[Feng et~al\mbox{.}(2023)]%
        {feng2023refinetad}
\bibfield{author}{\bibinfo{person}{Yue Feng}, \bibinfo{person}{Zhengye Zhang}, \bibinfo{person}{Rong Quan}, \bibinfo{person}{Limin Wang}, {and} \bibinfo{person}{Jie Qin}.} \bibinfo{year}{2023}\natexlab{}.
\newblock \showarticletitle{RefineTAD: Learning Proposal-free Refinement for Temporal Action Detection}. In \bibinfo{booktitle}{\emph{ACM Int. Conf. Multimedia}}. \bibinfo{pages}{135--143}.
\newblock


\bibitem[Ghadiyaram et~al\mbox{.}(2019)]%
        {ghadiyaram2019large}
\bibfield{author}{\bibinfo{person}{Deepti Ghadiyaram}, \bibinfo{person}{Du Tran}, {and} \bibinfo{person}{Dhruv Mahajan}.} \bibinfo{year}{2019}\natexlab{}.
\newblock \showarticletitle{Large-scale weakly-supervised pre-training for video action recognition}. In \bibinfo{booktitle}{\emph{IEEE Conf. Comput. Vis. Pattern Recog.}} \bibinfo{pages}{12046--12055}.
\newblock


\bibitem[Gothe et~al\mbox{.}(2024)]%
        {gothe2024s}
\bibfield{author}{\bibinfo{person}{Sourabh~Vasant Gothe}, \bibinfo{person}{Vibhav Agarwal}, \bibinfo{person}{Sourav Ghosh}, \bibinfo{person}{Jayesh~Rajkumar Vachhani}, \bibinfo{person}{Pranay Kashyap}, {and} \bibinfo{person}{Barath Raj~Kandur Raja}.} \bibinfo{year}{2024}\natexlab{}.
\newblock \showarticletitle{What's in the Flow? Exploiting Temporal Motion Cues for Unsupervised Generic Event Boundary Detection}. In \bibinfo{booktitle}{\emph{IEEE Winter Conf. App. Comput. Vis.}} \bibinfo{pages}{6941--6950}.
\newblock


\bibitem[Gothe et~al\mbox{.}(2023)]%
        {gothe2023self}
\bibfield{author}{\bibinfo{person}{Sourabh~Vasant Gothe}, \bibinfo{person}{Jayesh~Rajkumar Vachhani}, \bibinfo{person}{Rishabh Khurana}, {and} \bibinfo{person}{Pranay Kashyap}.} \bibinfo{year}{2023}\natexlab{}.
\newblock \showarticletitle{Self-Similarity is all You Need for Fast and Light-Weight Generic Event Boundary Detection}. In \bibinfo{booktitle}{\emph{ICASSP}}. IEEE, \bibinfo{pages}{1--5}.
\newblock


\bibitem[Gygli(2018)]%
        {gygli2018ridiculously}
\bibfield{author}{\bibinfo{person}{Michael Gygli}.} \bibinfo{year}{2018}\natexlab{}.
\newblock \showarticletitle{Ridiculously fast shot boundary detection with fully convolutional neural networks}. In \bibinfo{booktitle}{\emph{2018 International Conference on Content-Based Multimedia Indexing (CBMI)}}. IEEE, \bibinfo{pages}{1--4}.
\newblock


\bibitem[He et~al\mbox{.}(2016)]%
        {he2016deep}
\bibfield{author}{\bibinfo{person}{Kaiming He}, \bibinfo{person}{Xiangyu Zhang}, \bibinfo{person}{Shaoqing Ren}, {and} \bibinfo{person}{Jian Sun}.} \bibinfo{year}{2016}\natexlab{}.
\newblock \showarticletitle{Deep residual learning for image recognition}. In \bibinfo{booktitle}{\emph{IEEE Conf. Comput. Vis. Pattern Recog.}} \bibinfo{pages}{770--778}.
\newblock


\bibitem[He et~al\mbox{.}(2019)]%
        {he2019unsupervised}
\bibfield{author}{\bibinfo{person}{Xufeng He}, \bibinfo{person}{Yang Hua}, \bibinfo{person}{Tao Song}, \bibinfo{person}{Zongpu Zhang}, \bibinfo{person}{Zhengui Xue}, \bibinfo{person}{Ruhui Ma}, \bibinfo{person}{Neil Robertson}, {and} \bibinfo{person}{Haibing Guan}.} \bibinfo{year}{2019}\natexlab{}.
\newblock \showarticletitle{Unsupervised video summarization with attentive conditional generative adversarial networks}. In \bibinfo{booktitle}{\emph{ACM Int. Conf. Multimedia}}. \bibinfo{pages}{2296--2304}.
\newblock


\bibitem[Hong et~al\mbox{.}(2021)]%
        {hong2021generic}
\bibfield{author}{\bibinfo{person}{Dexiang Hong}, \bibinfo{person}{Congcong Li}, \bibinfo{person}{Longyin Wen}, \bibinfo{person}{Xinyao Wang}, {and} \bibinfo{person}{Libo Zhang}.} \bibinfo{year}{2021}\natexlab{}.
\newblock \showarticletitle{Generic event boundary detection challenge at CVPR 2021 technical report: Cascaded temporal attention network (CASTANET)}.
\newblock \bibinfo{journal}{\emph{arXiv preprint arXiv:2107.00239}} (\bibinfo{year}{2021}).
\newblock


\bibitem[Hong et~al\mbox{.}(2022)]%
        {hong2022sc}
\bibfield{author}{\bibinfo{person}{Dexiang Hong}, \bibinfo{person}{Xiaoqi Ma}, \bibinfo{person}{Xinyao Wang}, \bibinfo{person}{Congcong Li}, \bibinfo{person}{Yufei Wang}, {and} \bibinfo{person}{Longyin Wen}.} \bibinfo{year}{2022}\natexlab{}.
\newblock \showarticletitle{SC-Transformer++: Structured Context Transformer for Generic Event Boundary Detection}.
\newblock \bibinfo{journal}{\emph{arXiv preprint arXiv:2206.12634}} (\bibinfo{year}{2022}).
\newblock


\bibitem[Hu et~al\mbox{.}(2018)]%
        {hu2018squeeze}
\bibfield{author}{\bibinfo{person}{Jie Hu}, \bibinfo{person}{Li Shen}, {and} \bibinfo{person}{Gang Sun}.} \bibinfo{year}{2018}\natexlab{}.
\newblock \showarticletitle{Squeeze-and-excitation networks}. In \bibinfo{booktitle}{\emph{IEEE Conf. Comput. Vis. Pattern Recog.}} \bibinfo{pages}{7132--7141}.
\newblock


\bibitem[Huang et~al\mbox{.}(2016)]%
        {huang2016connectionist}
\bibfield{author}{\bibinfo{person}{De-An Huang}, \bibinfo{person}{Li Fei-Fei}, {and} \bibinfo{person}{Juan~Carlos Niebles}.} \bibinfo{year}{2016}\natexlab{}.
\newblock \showarticletitle{Connectionist temporal modeling for weakly supervised action labeling}. In \bibinfo{booktitle}{\emph{Eur. Conf. Comput. Vis.}} Springer, \bibinfo{pages}{137--153}.
\newblock


\bibitem[Huynh et~al\mbox{.}(2023)]%
        {huynh2023generic}
\bibfield{author}{\bibinfo{person}{Van~Thong Huynh}, \bibinfo{person}{Hyung-Jeong Yang}, \bibinfo{person}{Guee-Sang Lee}, {and} \bibinfo{person}{Soo-Hyung Kim}.} \bibinfo{year}{2023}\natexlab{}.
\newblock \showarticletitle{Generic Event Boundary Detection in Video with Pyramid Features}.
\newblock \bibinfo{journal}{\emph{arXiv preprint arXiv:2301.04288}} (\bibinfo{year}{2023}).
\newblock


\bibitem[Ioffe and Szegedy(2015)]%
        {ioffe2015batch}
\bibfield{author}{\bibinfo{person}{Sergey Ioffe} {and} \bibinfo{person}{Christian Szegedy}.} \bibinfo{year}{2015}\natexlab{}.
\newblock \showarticletitle{Batch normalization: Accelerating deep network training by reducing internal covariate shift}. In \bibinfo{booktitle}{\emph{Int. Conf. Machine Learn.}} \bibinfo{pages}{448--456}.
\newblock


\bibitem[Kang et~al\mbox{.}(2021)]%
        {kang2021winning}
\bibfield{author}{\bibinfo{person}{Hyolim Kang}, \bibinfo{person}{Jinwoo Kim}, \bibinfo{person}{Kyungmin Kim}, \bibinfo{person}{Taehyun Kim}, {and} \bibinfo{person}{Seon~Joo Kim}.} \bibinfo{year}{2021}\natexlab{}.
\newblock \showarticletitle{Winning the CVPR'2021 Kinetics-GEBD Challenge: Contrastive Learning Approach}.
\newblock \bibinfo{journal}{\emph{arXiv preprint arXiv:2106.11549}} (\bibinfo{year}{2021}).
\newblock


\bibitem[Kang et~al\mbox{.}(2022)]%
        {kang2022uboco}
\bibfield{author}{\bibinfo{person}{Hyolim Kang}, \bibinfo{person}{Jinwoo Kim}, \bibinfo{person}{Taehyun Kim}, {and} \bibinfo{person}{Seon~Joo Kim}.} \bibinfo{year}{2022}\natexlab{}.
\newblock \showarticletitle{Uboco: Unsupervised boundary contrastive learning for generic event boundary detection}. In \bibinfo{booktitle}{\emph{IEEE Conf. Comput. Vis. Pattern Recog.}} \bibinfo{pages}{20073--20082}.
\newblock


\bibitem[Kingma and Ba(2014)]%
        {kingma2014adam}
\bibfield{author}{\bibinfo{person}{Diederik~P Kingma} {and} \bibinfo{person}{Jimmy Ba}.} \bibinfo{year}{2014}\natexlab{}.
\newblock \showarticletitle{Adam: A method for stochastic optimization}.
\newblock \bibinfo{journal}{\emph{arXiv preprint arXiv:1412.6980}} (\bibinfo{year}{2014}).
\newblock


\bibitem[Lea et~al\mbox{.}(2016)]%
        {lea2016segmental}
\bibfield{author}{\bibinfo{person}{Colin Lea}, \bibinfo{person}{Austin Reiter}, \bibinfo{person}{Ren{\'e} Vidal}, {and} \bibinfo{person}{Gregory~D Hager}.} \bibinfo{year}{2016}\natexlab{}.
\newblock \showarticletitle{Segmental spatiotemporal cnns for fine-grained action segmentation}. In \bibinfo{booktitle}{\emph{Eur. Conf. Comput. Vis.}} Springer, \bibinfo{pages}{36--52}.
\newblock


\bibitem[Lei and Todorovic(2018)]%
        {lei2018temporal}
\bibfield{author}{\bibinfo{person}{Peng Lei} {and} \bibinfo{person}{Sinisa Todorovic}.} \bibinfo{year}{2018}\natexlab{}.
\newblock \showarticletitle{Temporal deformable residual networks for action segmentation in videos}. In \bibinfo{booktitle}{\emph{IEEE Conf. Comput. Vis. Pattern Recog.}} \bibinfo{pages}{6742--6751}.
\newblock


\bibitem[Li et~al\mbox{.}(2022a)]%
        {li2022structured}
\bibfield{author}{\bibinfo{person}{Congcong Li}, \bibinfo{person}{Xinyao Wang}, \bibinfo{person}{Dexiang Hong}, \bibinfo{person}{Yufei Wang}, \bibinfo{person}{Libo Zhang}, \bibinfo{person}{Tiejian Luo}, {and} \bibinfo{person}{Longyin Wen}.} \bibinfo{year}{2022}\natexlab{a}.
\newblock \showarticletitle{Structured context transformer for generic event boundary detection}.
\newblock \bibinfo{journal}{\emph{arXiv preprint arXiv:2206.02985}} (\bibinfo{year}{2022}).
\newblock


\bibitem[Li et~al\mbox{.}(2022b)]%
        {li2022end}
\bibfield{author}{\bibinfo{person}{Congcong Li}, \bibinfo{person}{Xinyao Wang}, \bibinfo{person}{Longyin Wen}, \bibinfo{person}{Dexiang Hong}, \bibinfo{person}{Tiejian Luo}, {and} \bibinfo{person}{Libo Zhang}.} \bibinfo{year}{2022}\natexlab{b}.
\newblock \showarticletitle{End-to-end compressed video representation learning for generic event boundary detection}. In \bibinfo{booktitle}{\emph{IEEE Conf. Comput. Vis. Pattern Recog.}} \bibinfo{pages}{13967--13976}.
\newblock


\bibitem[Li et~al\mbox{.}(2020)]%
        {li2020learning}
\bibfield{author}{\bibinfo{person}{Yanwei Li}, \bibinfo{person}{Lin Song}, \bibinfo{person}{Yukang Chen}, \bibinfo{person}{Zeming Li}, \bibinfo{person}{Xiangyu Zhang}, \bibinfo{person}{Xingang Wang}, {and} \bibinfo{person}{Jian Sun}.} \bibinfo{year}{2020}\natexlab{}.
\newblock \showarticletitle{Learning dynamic routing for semantic segmentation}. In \bibinfo{booktitle}{\emph{IEEE Conf. Comput. Vis. Pattern Recog.}} \bibinfo{pages}{8553--8562}.
\newblock


\bibitem[Lin et~al\mbox{.}(2019)]%
        {lin2019bmn}
\bibfield{author}{\bibinfo{person}{Tianwei Lin}, \bibinfo{person}{Xiao Liu}, \bibinfo{person}{Xin Li}, \bibinfo{person}{Errui Ding}, {and} \bibinfo{person}{Shilei Wen}.} \bibinfo{year}{2019}\natexlab{}.
\newblock \showarticletitle{Bmn: Boundary-matching network for temporal action proposal generation}. In \bibinfo{booktitle}{\emph{Int. Conf. Comput. Vis.}} \bibinfo{pages}{3889--3898}.
\newblock


\bibitem[Lin et~al\mbox{.}(2017a)]%
        {lin2017feature}
\bibfield{author}{\bibinfo{person}{Tsung-Yi Lin}, \bibinfo{person}{Piotr Doll{\'a}r}, \bibinfo{person}{Ross Girshick}, \bibinfo{person}{Kaiming He}, \bibinfo{person}{Bharath Hariharan}, {and} \bibinfo{person}{Serge Belongie}.} \bibinfo{year}{2017}\natexlab{a}.
\newblock \showarticletitle{Feature pyramid networks for object detection}. In \bibinfo{booktitle}{\emph{Proceedings of the IEEE conference on computer vision and pattern recognition}}. \bibinfo{pages}{2117--2125}.
\newblock


\bibitem[Lin et~al\mbox{.}(2017b)]%
        {lin2017focal}
\bibfield{author}{\bibinfo{person}{Tsung-Yi Lin}, \bibinfo{person}{Priya Goyal}, \bibinfo{person}{Ross Girshick}, \bibinfo{person}{Kaiming He}, {and} \bibinfo{person}{Piotr Doll{\'a}r}.} \bibinfo{year}{2017}\natexlab{b}.
\newblock \showarticletitle{Focal loss for dense object detection}. In \bibinfo{booktitle}{\emph{Int. Conf. Comput. Vis.}} \bibinfo{pages}{2980--2988}.
\newblock


\bibitem[Mizufune et~al\mbox{.}(2023)]%
        {mizufune2023margin}
\bibfield{author}{\bibinfo{person}{Kosuke Mizufune}, \bibinfo{person}{Shunsuke Tanaka}, \bibinfo{person}{Toshihide Yukitake}, {and} \bibinfo{person}{Tatsushi Matsubayashi}.} \bibinfo{year}{2023}\natexlab{}.
\newblock \showarticletitle{Margin MCC: Chance-Robust Metric for Video Boundary Detection with Allowed Margin}. In \bibinfo{booktitle}{\emph{ACM Int. Conf. Multimedia}}. \bibinfo{pages}{2694--2703}.
\newblock


\bibitem[Nag et~al\mbox{.}(2023)]%
        {nag2023post}
\bibfield{author}{\bibinfo{person}{Sauradip Nag}, \bibinfo{person}{Xiatian Zhu}, \bibinfo{person}{Yi-Zhe Song}, {and} \bibinfo{person}{Tao Xiang}.} \bibinfo{year}{2023}\natexlab{}.
\newblock \showarticletitle{Post-Processing Temporal Action Detection}. In \bibinfo{booktitle}{\emph{IEEE Conf. Comput. Vis. Pattern Recog.}} \bibinfo{pages}{18837--18845}.
\newblock


\bibitem[Radvansky and Zacks(2011)]%
        {radvansky2011event}
\bibfield{author}{\bibinfo{person}{Gabriel~A Radvansky} {and} \bibinfo{person}{Jeffrey~M Zacks}.} \bibinfo{year}{2011}\natexlab{}.
\newblock \showarticletitle{Event perception}.
\newblock \bibinfo{journal}{\emph{Wiley Interdisciplinary Reviews: Cognitive Science}} \bibinfo{volume}{2}, \bibinfo{number}{6} (\bibinfo{year}{2011}), \bibinfo{pages}{608--620}.
\newblock


\bibitem[Rai et~al\mbox{.}(2023)]%
        {rai2023motion}
\bibfield{author}{\bibinfo{person}{Ayush~K Rai}, \bibinfo{person}{Tarun Krishna}, \bibinfo{person}{Julia Dietlmeier}, \bibinfo{person}{Kevin McGuinness}, \bibinfo{person}{Alan~F Smeaton}, {and} \bibinfo{person}{Noel~E O’Connor}.} \bibinfo{year}{2023}\natexlab{}.
\newblock \showarticletitle{Motion aware self-supervision for generic event boundary detection}. In \bibinfo{booktitle}{\emph{IEEE Winter Conf. App. Comput. Vis.}} \bibinfo{pages}{2728--2739}.
\newblock


\bibitem[Shao et~al\mbox{.}(2020)]%
        {shao2020intra}
\bibfield{author}{\bibinfo{person}{Dian Shao}, \bibinfo{person}{Yue Zhao}, \bibinfo{person}{Bo Dai}, {and} \bibinfo{person}{Dahua Lin}.} \bibinfo{year}{2020}\natexlab{}.
\newblock \showarticletitle{Intra-and inter-action understanding via temporal action parsing}. In \bibinfo{booktitle}{\emph{IEEE Conf. Comput. Vis. Pattern Recog.}} \bibinfo{pages}{730--739}.
\newblock


\bibitem[Shi et~al\mbox{.}(2023)]%
        {shi2023tridet}
\bibfield{author}{\bibinfo{person}{Dingfeng Shi}, \bibinfo{person}{Yujie Zhong}, \bibinfo{person}{Qiong Cao}, \bibinfo{person}{Lin Ma}, \bibinfo{person}{Jia Li}, {and} \bibinfo{person}{Dacheng Tao}.} \bibinfo{year}{2023}\natexlab{}.
\newblock \showarticletitle{Tridet: Temporal action detection with relative boundary modeling}. In \bibinfo{booktitle}{\emph{IEEE Conf. Comput. Vis. Pattern Recog.}} \bibinfo{pages}{18857--18866}.
\newblock


\bibitem[Shou et~al\mbox{.}(2021b)]%
        {loveu21}
\bibfield{author}{\bibinfo{person}{Mike~Zheng Shou}, \bibinfo{person}{Stan LEI}, \bibinfo{person}{Linchao Zhu}, \bibinfo{person}{Xiaohan Wang}, {et~al\mbox{.}}} \bibinfo{year}{2021}\natexlab{b}.
\newblock \bibinfo{booktitle}{}.
\newblock
\urldef\tempurl%
\url{https://sites.google.com/view/loveucvpr21/track-1}
\showURL{%
\tempurl}


\bibitem[Shou et~al\mbox{.}(2021a)]%
        {shou2021generic}
\bibfield{author}{\bibinfo{person}{Mike~Zheng Shou}, \bibinfo{person}{Stan~Weixian Lei}, \bibinfo{person}{Weiyao Wang}, \bibinfo{person}{Deepti Ghadiyaram}, {and} \bibinfo{person}{Matt Feiszli}.} \bibinfo{year}{2021}\natexlab{a}.
\newblock \showarticletitle{Generic event boundary detection: A benchmark for event segmentation}. In \bibinfo{booktitle}{\emph{Int. Conf. Comput. Vis.}} \bibinfo{pages}{8075--8084}.
\newblock


\bibitem[Shou et~al\mbox{.}(2022)]%
        {loveu22}
\bibfield{author}{\bibinfo{person}{Mike~Zheng Shou}, \bibinfo{person}{Linchao Zhu}, \bibinfo{person}{Lorenzo Torresani}, \bibinfo{person}{Kristen Grauman}, \bibinfo{person}{Matt Feiszli}, {et~al\mbox{.}}} \bibinfo{year}{2022}\natexlab{}.
\newblock \bibinfo{booktitle}{}.
\newblock
\urldef\tempurl%
\url{https://sites.google.com/view/loveucvpr22/track-1}
\showURL{%
\tempurl}


\bibitem[Shou et~al\mbox{.}(2023)]%
        {loveu23}
\bibfield{author}{\bibinfo{person}{Mike~Zheng Shou}, \bibinfo{person}{Linchao Zhu}, \bibinfo{person}{Lorenzo Torresani}, \bibinfo{person}{Kristen Grauman}, \bibinfo{person}{Matt Feiszli}, {et~al\mbox{.}}} \bibinfo{year}{2023}\natexlab{}.
\newblock \bibinfo{booktitle}{}.
\newblock
\urldef\tempurl%
\url{https://sites.google.com/view/loveucvpr23/track1}
\showURL{%
\tempurl}


\bibitem[Sou{\v{c}}ek et~al\mbox{.}(2019)]%
        {souvcek2019transnet}
\bibfield{author}{\bibinfo{person}{Tom{\'a}{\v{s}} Sou{\v{c}}ek}, \bibinfo{person}{Jaroslav Moravec}, {and} \bibinfo{person}{Jakub Loko{\v{c}}}.} \bibinfo{year}{2019}\natexlab{}.
\newblock \showarticletitle{Transnet: A deep network for fast detection of common shot transitions}.
\newblock \bibinfo{journal}{\emph{arXiv preprint arXiv:1906.03363}} (\bibinfo{year}{2019}).
\newblock


\bibitem[Sun et~al\mbox{.}(2023)]%
        {sun2023mae}
\bibfield{author}{\bibinfo{person}{Yuanxi Sun}, \bibinfo{person}{Rui He}, \bibinfo{person}{Youzeng Li}, \bibinfo{person}{Zuwei Huang}, \bibinfo{person}{Feng Hu}, \bibinfo{person}{Xu Cheng}, {and} \bibinfo{person}{Jie Tang}.} \bibinfo{year}{2023}\natexlab{}.
\newblock \showarticletitle{MAE-GEBD: Winning the CVPR'2023 LOVEU-GEBD Challenge}.
\newblock \bibinfo{journal}{\emph{arXiv preprint arXiv:2306.15704}} (\bibinfo{year}{2023}).
\newblock


\bibitem[Tan et~al\mbox{.}(2023)]%
        {tan2023temporal}
\bibfield{author}{\bibinfo{person}{Jing Tan}, \bibinfo{person}{Yuhong Wang}, \bibinfo{person}{Gangshan Wu}, {and} \bibinfo{person}{Limin Wang}.} \bibinfo{year}{2023}\natexlab{}.
\newblock \showarticletitle{Temporal Perceiver: A General Architecture for Arbitrary Boundary Detection}.
\newblock \bibinfo{journal}{\emph{IEEE Trans. Pattern Anal. Mach. Intell.}} (\bibinfo{year}{2023}).
\newblock


\bibitem[Tang et~al\mbox{.}(2022)]%
        {tang2022progressive}
\bibfield{author}{\bibinfo{person}{Jiaqi Tang}, \bibinfo{person}{Zhaoyang Liu}, \bibinfo{person}{Chen Qian}, \bibinfo{person}{Wayne Wu}, {and} \bibinfo{person}{Limin Wang}.} \bibinfo{year}{2022}\natexlab{}.
\newblock \showarticletitle{Progressive attention on multi-level dense difference maps for generic event boundary detection}. In \bibinfo{booktitle}{\emph{IEEE Conf. Comput. Vis. Pattern Recog.}} \bibinfo{pages}{3355--3364}.
\newblock


\bibitem[Tang et~al\mbox{.}(2018)]%
        {tang2018fast}
\bibfield{author}{\bibinfo{person}{Shitao Tang}, \bibinfo{person}{Litong Feng}, \bibinfo{person}{Zhanghui Kuang}, \bibinfo{person}{Yimin Chen}, {and} \bibinfo{person}{Wei Zhang}.} \bibinfo{year}{2018}\natexlab{}.
\newblock \showarticletitle{Fast video shot transition localization with deep structured models}. In \bibinfo{booktitle}{\emph{Asian Conference on Computer Vision}}. Springer, \bibinfo{pages}{577--592}.
\newblock


\bibitem[Tran et~al\mbox{.}(2019)]%
        {tran2019video}
\bibfield{author}{\bibinfo{person}{Du Tran}, \bibinfo{person}{Heng Wang}, \bibinfo{person}{Lorenzo Torresani}, {and} \bibinfo{person}{Matt Feiszli}.} \bibinfo{year}{2019}\natexlab{}.
\newblock \showarticletitle{Video classification with channel-separated convolutional networks}. In \bibinfo{booktitle}{\emph{Int. Conf. Comput. Vis.}} \bibinfo{pages}{5552--5561}.
\newblock


\bibitem[Wu et~al\mbox{.}(2018)]%
        {wu2018compressed}
\bibfield{author}{\bibinfo{person}{Chao-Yuan Wu}, \bibinfo{person}{Manzil Zaheer}, \bibinfo{person}{Hexiang Hu}, \bibinfo{person}{R Manmatha}, \bibinfo{person}{Alexander~J Smola}, {and} \bibinfo{person}{Philipp Kr{\"a}henb{\"u}hl}.} \bibinfo{year}{2018}\natexlab{}.
\newblock \showarticletitle{Compressed video action recognition}. In \bibinfo{booktitle}{\emph{IEEE Conf. Comput. Vis. Pattern Recog.}} \bibinfo{pages}{6026--6035}.
\newblock


\bibitem[Zhang et~al\mbox{.}(2022)]%
        {zhang2022actionformer}
\bibfield{author}{\bibinfo{person}{Chen-Lin Zhang}, \bibinfo{person}{Jianxin Wu}, {and} \bibinfo{person}{Yin Li}.} \bibinfo{year}{2022}\natexlab{}.
\newblock \showarticletitle{Actionformer: Localizing moments of actions with transformers}. In \bibinfo{booktitle}{\emph{Eur. Conf. Comput. Vis.}} Springer, \bibinfo{pages}{492--510}.
\newblock


\bibitem[Zhang et~al\mbox{.}(2023)]%
        {zhang2023local}
\bibfield{author}{\bibinfo{person}{Libo Zhang}, \bibinfo{person}{Xin Gu}, \bibinfo{person}{Congcong Li}, \bibinfo{person}{Tiejian Luo}, {and} \bibinfo{person}{Heng Fan}.} \bibinfo{year}{2023}\natexlab{}.
\newblock \showarticletitle{Local Compressed Video Stream Learning for Generic Event Boundary Detection}.
\newblock \bibinfo{journal}{\emph{Int. J. Comput. Vis.}} (\bibinfo{year}{2023}), \bibinfo{pages}{1--18}.
\newblock


\bibitem[Zheng et~al\mbox{.}(2019)]%
        {zheng2019relation}
\bibfield{author}{\bibinfo{person}{Sipeng Zheng}, \bibinfo{person}{Xiangyu Chen}, \bibinfo{person}{Shizhe Chen}, {and} \bibinfo{person}{Qin Jin}.} \bibinfo{year}{2019}\natexlab{}.
\newblock \showarticletitle{Relation understanding in videos}. In \bibinfo{booktitle}{\emph{ACM Int. Conf. Multimedia}}. \bibinfo{pages}{2662--2666}.
\newblock


\bibitem[Zheng et~al\mbox{.}(2024)]%
        {zheng2023dynamic}
\bibfield{author}{\bibinfo{person}{Ziwei Zheng}, \bibinfo{person}{Le Yang}, \bibinfo{person}{Yulin Wang}, \bibinfo{person}{Miao Zhang}, \bibinfo{person}{Lijun He}, \bibinfo{person}{Gao Huang}, {and} \bibinfo{person}{Fan Li}.} \bibinfo{year}{2024}\natexlab{}.
\newblock \showarticletitle{Dynamic Spatial Focus for Efficient Compressed Video Action Recognition}.
\newblock \bibinfo{journal}{\emph{IEEE Trans. Circuit Syst. Video Technol.}} (\bibinfo{year}{2024}).
\newblock


\end{thebibliography}










\end{document}


\title{Supplementary Materials: Rethinking the Architecture Design for Efficient Generic Event Boundary Detection}


\author{Anonymous Authors}

\maketitle

\section{Architecture details and training settings}

\section{Fully experimental results}

\begin{table}[t]
\centering
\caption{The architectures of three representative GEBD methods and the propose models in this paper. }
\label{tab:base}
\resizebox{1.01\linewidth}{!}{
\begin{tabular}{llll}
\toprule
Method	&	F1@0.05	&	GFLOPs	&	FPS	\\ \midrule
SC-Transformer	&		&		&		&		&		\\
DDM-Net	&		&		&		&		&		\\ \midrule
\namebasic{}	&	77.1	&		&		&	4.36	&	1562	\\
\rowcolor{backbone} Res50-L3	&	77	&		&		&	3.57	&	1783	\\
\rowcolor{backbone}\namebasic{} (Res50-L2)	&	76.8	&		&		&	2.08	&	2325	\\
\rowcolor{backbone}\namebasic{} (Res50-L1)	&	75.3	&		&		&	1.05	&	2699	\\
\rowcolor{backbone}\namebasic{} (Res34-L4) &	0.7702	&	0.7381	&	0.8051	&	3.92	&	2386	\\
\rowcolor{backbone}\namebasic{} (Res34-L2) &	0.766	&	0.7318	&	0.8035	&	1.94	&	2495	\\
\rowcolor{backbone}\namebasic{} (Res18-L2)	&	76.1	&	71.7	&	81.2	&	1.24	&	2480	\\
\rowcolor{backbone}\namebasic{} (Res18-L4)	&	77.2	&	74.3	&	80.4	&	2.07	&	2380	\\
\rowcolor{backbone}\namebasic{} (Res152-L2)	&	76.3	&	73.5	&	79.3	&	2.94	&	1783	\\
\rowcolor{backbone}\namebasic{} (Res152-L4)	&	77.2	&	74.5	&	80.1	&	11.77	&	847	\\ \midrule

\rowcolor{encoder} 1d-conv	&	76.4	&		&		&	2.09	&	2257	\\
\rowcolor{encoder}Diff Mixer x1	&	76.5	&		&		&		&		\\
\rowcolor{encoder}Diff former x1	&	76.3	&		&		&		&		\\
\rowcolor{encoder}Diff Mixer x2	&	76.5	&		&		&		&		\\
\rowcolor{encoder}Diff former x2	&	76.2	&		&		&		&		\\ \midrule

\rowcolor{fuse} Cat. fuse	&	77.6	&		&		&		&	2208	\\
\rowcolor{fuse} Cross Att.	&	77.7	&		&		&		&		\\ \midrule
\rowcolor{fcn} no 2d-FCN	&	64.4	&		&		&	1.85	&	2426	\\
\rowcolor{fcn} FCN-Res10	&	77.7	&		&		&	2.09	&	2208	\\
\rowcolor{fcn} FCN-Res10	&	77.8	&		&		&	2.42	&	2181	\\ \midrule
\nameshort{} (Res18-L2*)&	77.3	&	77.4	&	77.1	&	1.28	&	2477	\\
\nameshort{} (Res18-L4*)	&	78.2	&	78.7	&	77.8	&	2.11	&	2567	\\
\nameshort{} (Res34-L2*)	&	77.8	&	77.2	&	78.5	&	1.97	&	2529	\\
\nameshort{} (Res34-L4*)	&	78.5	&	78.4	&	78.6	&	3.96	&	2455	\\
\nameshort{} (Res50-L2*) \\
\rowcolor{nouse}\nameshort{} (Res50-L4*)   &&& &&\\
\rowcolor{nouse} \nameshort{} (Res152-L2*)	&	77.9	&	77.4	&	78.4	&	2.97	&	1685	\\
\rowcolor{nouse} \nameshort{} (Res152-L4*)	&	78.9	&	78.3	&	79.4	&	11.80	&	818	\\ \midrule


\rowcolor{video}  \nameshort{} (CSNR152-L2)&	80.4	&	80.1	&	80.6	&	1.98	&	2029	\\
\rowcolor{video}  \nameshort{} (CSNR152-L4)	&	82	&	81.2	&	82.7	&	6.37	&	1050	\\
\rowcolor{video}  \nameshort{} (CSNR152-L2*)	&	80.6	&	80	&	81.3	&	2.00	&	1215	\\
\rowcolor{video}  \nameshort{} (CSNR152-L4*)	&	82.3	&	81.6	&	83.01	&	6.40	&	1025	\\
\rowcolor{video} \namebasic{} (CSNR50-L4)	&		\\
\rowcolor{video} \namebasic{} (CSNR152-L4)	&	82.5	&	81.4	&	83.6	&	6.36	&	1054	\\

\bottomrule
\end{tabular}}
\end{table}

The title, subtitle, keywords and abstract will be typeset in the main
language of the paper.  The commands \verb|\translatedXXX|, \verb|XXX|
begin title, subtitle and keywords, can be used to set these elements
in the other languages.  The environment \verb|translatedabstract| is
used to set the translation of the abstract.  These commands and
environment have a mandatory first argument: the language of the
second argument.  See \verb|sample-sigconf-i13n.tex| file for examples
of their usage.

\section{Visualization of similarity maps}

\section{Visualization of  GradCAM++}

\section{exp results}

\begin{table*}[t]
\centering
\caption{Comparisons in terms o F1 score (\%) on Kinetics-GEBD with Rel.Dis. threshold from 0.05 to 0.5. 
}
\label{tab:exp_kgebd}
\resizebox{\textwidth}{!}{
\begin{tabular}{l|l|c|cccccccccc}
\toprule
\multirow{2}{*}{Method} & \multirow{2}{*}{Backbone} & \multicolumn{11}{c}{F1 @ Rel. Dis.}   \\ \cline{3-13} 
                    &    & 0.05          & 0.1           & 0.15          & 0.2           & 0.25          & 0.3           & 0.35          & 0.4           & 0.45          & 0.5           & avg           \\ \midrule
BMN \cite{lin2019bmn}         &            & 18.6          & 20.4          & 21.3          & 22.0          & 22.6          & 23.0          & 23.3          & 23.7          & 23.9          & 24.1          & 22.3          \\
BMN-StartEnd \cite{lin2019bmn}     &       & 49.1          & 58.9          & 62.7          & 64.8          & 66.0          & 66.8          & 67.4          & 67.8          & 68.1          & 68.3          & 64.0          \\
TCN-TAPOS \cite{lea2016segmental}     &          & 46.4          & 56.0          & 60.2          & 62.8          & 64.5          & 65.9          & 66.9          & 67.6          & 68.2          & 68.7          & 62.7          \\
TCN \cite{lea2016segmental}         &            & 58.8          & 65.7          & 67.9          & 69.1          & 69.8          & 70.3          & 70.6          & 70.8          & 71.0          & 71.2          & 68.5          \\
PC \cite{shou2021generic}          &            & 62.5          & 75.8          & 80.4          & 82.9          & 84.4          & 85.3          & 85.9          & 86.4          & 86.7          & 87.0          & 81.7          \\
SBoCo \cite{kang2022uboco}      &    ResNet50   & 73.2          & -             & -             & -             & -             & -             & -             & -             & -             & -             & 86.6          \\
Temporal Perceiver \cite{tan2023temporal}  &    & 74.8          & 82.8          & 85.2          & 86.6          & 87.4          & 87.9          & 88.3          & 88.7          & 89.0          & 89.2          & 86.0          \\
CVRL \cite{li2022end}        &            & 74.3          & 83.0          & 85.7          & 87.2          & 88.0          & 88.6          & 89.0          & 89.3          & 89.6          & 89.8          & 86.5          \\
CVRL+ \cite{zhang2023local}  & &76.8  &84.8 &87.2 &88.5 &89.2 &89.6 &89.9 &90.1 &90.3 &90.6 & 87.7\\ 
DDM-Net \cite{tang2022progressive}        &         & 76.4          & 84.3          & 86.6          & 88.0          & 88.7          & 89.2          & 89.5          & 89.8          & 90.0          & 90.2          & 87.3          \\
SC-Transformer \cite{li2022structured}   &       & 77.7          & 84.9          & 87.3          & 88.6          & 89.5          & 90.0          & 90.4          & 90.7          & 90.9          & 91.1          & 88.1          \\ 
\rowcolor{mygray} \textbf{\namebasic{}} &  ResNet50 & 77.2  &83.7   &86.0   &87.3   &88.1     &88.6   &89.0   &89.3   &89.5   &89.7   &86.9\\
 \textbf{\nameshort{}} &  ResNet50 & 78.3        & 85.1       & 87.4   & 88.7
  & 89.6    & 90.1    & 90.5   & 90.8  & 91.1  & 91.3   & 88.3
\\\midrule

Temporal Perceiver \cite{tan2023temporal}  &  CSN &82.2  &88.0  &89.9  &90.9  &91.6 &92.0   &92.3   &92.5   &92.7   &92.9   &90.5\\
CVRL+ \cite{zhang2023local} & CSN  &81.2 &- &- &- &- &- &- &- &- &- &81.2\\
\rowcolor{mygray}\textbf{\nameshort{}} & CSN  & 80.6   & 86.6    & 88.6     & 89.9    & 90.7   & 91.1   & 91.5  & 91.8 & 92.1  & 92.3   & 89.5    \\
\rowcolor{mygray}\textbf{\nameshort{}} & CSN  & 82.3      & 87.9      & 89.7      & 90.9    & 91.5      & 92.0   & 92.3    & 92.6      & 92.8    & 93.0 & 90.5 \\
\bottomrule
\end{tabular}
}
\end{table*}

\begin{table*}[t]
\setlength{\tabcolsep}{5pt}
\centering
\caption{Comparison with others in terms of F1 score (\%) on TAPOS with Rel.Dis. threshold from 0.05 to 0.5 with 0.05 interval.}
\label{tab:exp_tapos}
\resizebox{\textwidth}{!}{
\begin{tabular}{l|c|cccccccccc}
\toprule
\multirow{2}{*}{Method} & \multicolumn{11}{c}{F1 @ Rel. Dis.}                                                                                                                                                                                                                                                                    \\ \cline{2-12} 
                        & 0.05                     & 0.1                      & 0.15                     & 0.2                      & 0.25                     & 0.3                      & 0.35                     & 0.4                      & 0.45                     & 0.5                      & avg                      \\ \midrule
ISBA \cite{ding2018weakly}                   & 10.6                     & 17.0                     & 22.7                     & 26.5                     & 29.8                     & 32.6                     & 34.8                     & 36.9                     & 38.2                     & 39.6                     & 30.2                     \\
TCN \cite{lea2016segmental}                    & 23.7                     & 31.2                     & 33.1                     & 33.9                     & 34.2                     & 34.4                     & 34.7                     & 34.8                     & 34.8                     & 34.8                     & 64.0                     \\
CTM \cite{huang2016connectionist}                    & 24.4                     & 31.2                     & 33.6                     & 35.1                     & 36.1                     & 36.9                     & 37.4                     & 38.1                     & 38.3                     & 38.5                     & 35.0                     \\
TransParser \cite{shao2020intra}            & 28.9                     & 38.1                     & 43.5                     & 47.5                     & 50.0                     & 51.4                     & 52.7                     & 53.4                     & 54.0                     & 54.5                     & 47.4                     \\
PC \cite{shou2021generic}                     & 52.2                     & 59.5                     & 62.8                     & 64.6                     & 65.9                     & 66.5                     & 67.1                     & 67.6                     & 67.9                     & 68.3                     & 64.2                     \\
Temporal Perceiver \cite{tan2023temporal}     & 55.2 & 66.3 & 71.3 & 73.8 & 75.7 & 76.5 & 77.4 & 77.9 & 78.4 & 78.8 & 73.2 \\
DDM-Net \cite{tang2022progressive}                & 60.4                     & 68.1                     & 71.5                     & 73.5                     & 74.7                     & 75.3                     & 75.7                     & 76.0                     & 76.3                     & 76.7                     & 72.8                     \\
SC-Transformer \cite{li2022structured}         & 61.8 & 69.4 & 72.8 & 74.9 & 76.1 & 76.7 & 77.1 & 77.4 & 77.7 & 78.0 & 74.2 \\ \midrule

\rowcolor{mygray} \textbf{DyBDet (ours)}  & \textbf{62.5}            & \textbf{70.1}            & \textbf{73.4}            & \textbf{75.6}            & \textbf{76.7}            & \textbf{77.2}            & \textbf{77.5}            & \textbf{77.9}            & \textbf{78.1}            & \textbf{78.4}            & \textbf{74.7}            \\ \bottomrule
\end{tabular}
}
\end{table*}

\section{More visualizations}


\bibliographystyle{ACM-Reference-Format}
\bibliography{sample-base}








